\documentclass[fleqn,10pt]{article}
\usepackage[a4paper, total={7in, 10in}]{geometry}

\newcommand{\figref}[1]{{Fig.~\ref{#1}}}

\newcommand{\supref}[1]{{\hyperref[supplementary]{Supplementary~{#1}}}}
\usepackage{lineno}
\usepackage[font=small]{caption}
\usepackage{graphicx}
\usepackage{subcaption}
\usepackage{comment}
\usepackage[utf8]{inputenc}
\usepackage[T1]{fontenc}
\usepackage{bm}
\usepackage{hyperref}
\usepackage{authblk}
\usepackage{xcolor}
\usepackage{amsmath}
\usepackage{amsfonts}
\usepackage{moreverb} 
\usepackage{algorithm}
\usepackage{algpseudocode}
\usepackage{paralist}
\usepackage{array}
\usepackage{booktabs}
\usepackage{makecell}
\usepackage{multirow}
\newcommand{\algoref}[1]{{Alg.~\ref{#1}}}
\newcommand\CONDITION[2]%
   {\begin{tabular}[t]{@{}l@{}}#1#2\end{tabular}}
\algdef{SE}[WHILE]{While}{EndWhile}[1]%
   {\algorithmicwhile\ \CONDITION{#1}{\ \algorithmicdo}}%
   {\algorithmicend\ \algorithmicwhile}
\algdef{SE}[FOR]{For}{EndFor}[1]%
   {\algorithmicfor\ \CONDITION{#1}{\ \algorithmicdo}}%
   {\algorithmicend\ \algorithmicfor}
\algdef{S}[FOR]{ForAll}[1]%
   {\algorithmicforall\ \CONDITION{#1}{\ \algorithmicdo}}
\algdef{SE}[REPEAT]{Repeat}{Until}%
   {\algorithmicrepeat}[1]%
   {\algorithmicuntil\ \CONDITION{#1}{}}
\algdef{SE}[IF]{If}{EndIf}[1]%
   {\algorithmicif\ \CONDITION{#1}{\ \algorithmicthen}}
   {\algorithmicend\ \algorithmicif}
\algdef{C}[IF]{IF}{ElsIf}[1]%
   {\algorithmicelse\ \algorithmicif\ \CONDITION{#1}{\ \algorithmicthen}}
\makeatletter
\ifthenelse{\equal{\ALG@noend}{t}}%
   {\algtext*{EndWhile}%
    \algtext*{EndFor}%
    \algtext*{EndLoop}%
    \algtext*{EndIf}%
   }{}%
\makeatother

\makeatletter
\algnewcommand{\LineComment}[1]{\Statex \hskip\ALG@thistlm \(\triangleright\) #1}
\makeatother

\usepackage{xr}
\makeatletter
\newcommand*{\addFileDependency}[1]{
\typeout{(#1)}
\@addtofilelist{#1}
\IfFileExists{#1}{}{\typeout{No file #1.}}
}\makeatother

\immediate\write18{texcount -inc -incbib 
-sum main.tex > /tmp/wordcount.tex}

\title{DCAST: Diverse Class-Aware Self-Training Mitigates Selection Bias for Fairer Learning}

\author[1]{Yasin I. Tepeli}
\author[1,*]{Joana P. Gonçalves}
\affil[1]{Department of Intelligent Systems, Faculty EEMCS, Delft, Netherlands}

\affil[*]{Correspondence: joana.goncalves@tudelft.nl}

\begin{document}
\maketitle
\begin{abstract}
Fairness in machine learning seeks to mitigate model bias against individuals based on sensitive features such as sex or age, often caused by an uneven representation of the population in the training data due to selection bias. Notably, bias unascribed to sensitive features is challenging to identify and typically goes undiagnosed, despite its prominence in complex high-dimensional data from fields like computer vision and molecular biomedicine. Strategies to mitigate unidentified bias and evaluate mitigation methods are crucially needed, yet remain underexplored. We introduce: (i) Diverse Class-Aware Self-Training (DCAST), model-agnostic mitigation aware of class-specific bias, which promotes sample diversity to counter confirmation bias of conventional self-training while leveraging unlabeled samples for an improved representation of the underlying population; (ii) hierarchy bias, multivariate and class-aware bias induction without prior knowledge. Models learned with DCAST showed improved robustness to hierarchy and other biases across eleven datasets, against conventional self-training and six prominent domain adaptation techniques. Advantage was largest on multi-class classification, emphasizing DCAST as a promising strategy for fairer learning in different contexts.
\end{abstract}
\flushbottom
\maketitle

\thispagestyle{empty}
\section*{Introduction}
As predictive machine learning (ML) increasingly makes its way to applications with an impact on society, one major concern is to ensure that ML models deliver fair predictions and do not discriminate against individuals in the population. Selection bias is one of the most prominent sources of unfairness in ML, whereby the data used to build ML models is not representative of the real-world and thus violates the fundamental assumption of ML that it is independently drawn and identically distributed to the underlying population.

Research on fairness in ML has focused on mitigating (selection) bias associated with legally protected or sensitive features, such as sex, age, or skin color~\cite{Mehrabi2021, Pessach2022}. However, biases can be indirectly linked to sensitive features via proxies not recognized as sensitive~\cite{Mehrabi2021, Pessach2022}, or they can be unrelated to sensitive features and still lead to unfairness. 
Ultimately, biases are likely to remain undiagnosed and be propagated by ML models without scrutiny when a link to sensitive features is challenging to identify. Unknown biases are often present when data is complex and high-dimensional, data collection is non-random, and knowledge of the domain is incomplete. We argue that unfairness mitigation should thus address bias more generally, beyond what can be ascribed to sensitive features. 
This issue has deserved attention across fields, including computer vision~\cite{Wu2008, Persello2014}, astronomy~\cite{Richards2011, Kremer2015}, biomedicine and healthcare~\cite{Romero2011,Chan2020,Seale2022,Tepeli2024-jj}, finance and economics~\cite{Chang2009, Castagnetti2020, Shen2022}, information retrieval~\cite{Melucci2014, Melucci2016}, and language~\cite{Romero2011, Zhang2019_bias}. Nevertheless, its impact  is typically overlooked, resulting in models with optimistic performances due to bias-unaware evaluation. We identify two key areas for improvement, namely evaluation of ML model robustness to bias, and ML bias mitigation.

Evaluation is crucial to ensure that ML models generalize and are robust to bias, but assessing performance on data representative of the real-world distribution is rarely achievable. Independent test data is not always available or guaranteed to be unbiased, and conventional data splits do not create train-test distribution shifts suitable for model bias evaluation. A viable alternative is to induce bias to the train set and assess the learned model on the original test set. 
Common bias induction approaches include subsampling using univariate selection probabilities, based on values or the distribution of one feature~\cite{Chawla2005, Smith2007}. This is however not representative of multivariate biases typically present in complex high-dimensional data. Existing methods to induce multivariate bias include:  joint bias \cite{Huang2006}, which favors the selection of samples closer to the mean; and Dirichlet bias, \cite{Liu2014}, which assigns sample selection likelihoods based on a Dirichlet distribution. Both methods ignore class labels and thus do not generate class-specific biases. They might also cause class imbalances for otherwise balanced data. 

We propose hierarchy bias, a multivariate class-aware bias induction technique to produce complex class-specific biases. 
Hierarchy bias identifies distinctly distributed groups of samples in the original data using clustering, and then generates a biased selection by influencing the representation of one group of samples relative to the others. Selection is performed per class to induce class-specific bias, aiming for an identical number of samples per class to ensure class balance.

Methods to mitigate bias in ML generally fall in the scope of domain adaptation (DA,~\cite{Kouw2021review}), seeking to adapt a model to the distribution shift between the source training domain and a target prediction domain. Relevant DA categories span importance weighting, subspace alignment, inference-based, and semi-supervised learning methods. Importance weighting (IW) weighs training samples based on their relevance to the test set, using probability ratios or discrepancy measures ~\cite{Shimodaira2000, Zadrozny2004, Chang2009, Seah2011, Sugiyama2013, Kremer2015, Shen2018, Diesendruck2020, Huang2006, Du2021}. Since IW assumes that the train set contains the support of the test set and most features contribute to the prediction, it can be less effective with high-dimensional data or small sample sizes. Subspace alignment (SA) transforms the data representation~\cite{Blitzer2006, Fernando2013, Kouw2016}, assuming there is a common subspace where transformed train and test sets exhibit matching conditional probabilities, which may be difficult to optimize if many transformations fit. Inference-based (IB) methods include minimax estimation~\cite{Liu2014,Kouw2021tcpr}, where loss minimization is coupled with an adversarial maximization objective that steers the model to fit more conservatively, aiming for improved generalization. The IB methods may underperform if the model choice is less suitable for the test set. Overall, most IW, SA, and IB methods adapt the model for one target test set, which can hamper generalizability.
Semi-supervised learning (SSL) leverages unlabeled samples to provide model learning with insight into the underlying population distribution. The most benefit can in principle be achieved by using as much unlabeled data as available, though some SSL approaches still adapt to individual test sets~\cite{Fan2006, Ren2008}. Unlabeled samples are typically incorporated by SSL using self-training (ST)~\cite{McLachlan1975} or co-training (CT)~\cite{Blum1998}, which assigns predicted pseudo-labels to unlabeled samples and selects a subset of these to include at each training iteration. Sample selection is often based on prediction confidence according to the model trained thus far, which may strengthen existing bias or create other biases such as class imbalance for originally balanced data~\cite{Persello2014, Richards2011}. Attempts to mitigate this behavior include, for instance, the P3SVM support vector machine (SVM)~\cite{Persello2014} that selects pseudo-labeled samples distant from each other and located within the margins furthest away from the decision boundary. This method is however SVM-specific, and its sample selection dependent on the size of the margin may limit the contribution of unlabeled data. 
In summary, most DA methods mitigate distribution shifts for one test set at a time, leading to ML models with limited generalizability beyond the train and test domains. It remains to be investigated if generalization could be improved by training on additional unlabeled data.
Semi-supervised learning offers this possibility, but existing methods fall short in actively mitigating bias present in the data or further induced during model learning. Finally, many DA methods are model-specific and cannot be applied to different types of ML models. 

To improve bias mitigation, we propose Diverse Class-Aware Self-Training (DCAST), a model-agnostic semi-supervised learning framework that gradually incorporates unlabeled data in a class-aware manner, guided by two active bias mitigation strategies. 
The core CAST strategy addresses class-specific bias by selecting a set of pseudo-labeled samples to include separately per class, using a relaxed confidence threshold, with options to preserve the class ratios of the original labeled train set or to add the same number of pseudo-labeled samples per class at each iteration. The extended DCAST strategy seeks to counter confidence-induced bias by further selecting diverse pseudo-labeled samples, as measured by inter-sample distances in the learned discriminative embedding or the original feature space.  

We evaluate both hierarchy bias induction and (D)CAST bias mitigation across eleven datasets, against competing approaches including Dirichlet and joint bias as well as conventional self-training and six domain adaptation techniques. Specifically, we investigate which bias induction method induces the most challenging type of selection bias, leading to the strongest impact on ML model prediction performance. We further assess to what extent the class-awareness and diversity in (D)CAST improve robustness to bias, both across datasets and compared to the alternative bias mitigation strategies, while coupling model-agnostic (D)CAST with three types of ML models.

\section*{Results and Discussion}

The proposed hierarchy bias induction and (D)CAST bias mitigation methods aim to provide, respectively: (i) a more realistic type of class-aware multivariate selection bias for the evaluation of ML model robustness to bias, and (ii) class-aware and diversity-guided strategies to learn ML models with improved generalizability in the presence of selection bias. We briefly introduce these techniques and discuss their evaluation across 11 datasets using logistic regression (LR), random forest (RF), and 2-hidden layer neural network (NN) prediction models. Every dataset was randomly partitioned into 80\% train and 20\% test, with the test data reserved for prediction model evaluation (\hyperref[sec:methods_experimentalsettings]{Methods}). Effects of bias induction on the data and model prediction performance were assessed over 30 runs, each relying on a random split of the train set into labeled (30\%) and unlabeled (70\%) train sets. The labeled train set was used for bias induction and for training ML models, either intact or upon bias induction. For bias mitigation, unlabeled data was additionally used during training, where conventional self-training (ST) and (D)CAST leveraged the unlabeled train set, and other domain adaptation techniques exploited the unlabeled test set instead (\hyperref[sec:methods_experimentalsettings]{Methods}). 

\subsection*{Hierarchy bias induces effective multivariate and class-specific selection bias}
\begin{figure}[tbh]
\centering
\includegraphics[width=0.95\linewidth]{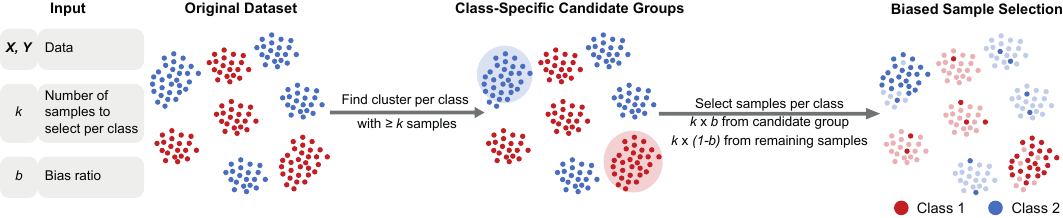}
\caption{\textbf{Hierarchy bias approach for induction of selection bias.} 
Given input data $\boldsymbol{X}$ with labels $\boldsymbol{Y}$, number of samples to select $k$, and bias ratio $b \in [0,1]$, hierarchy bias selects $k$ samples per class $c$: $k \times b$ from a specific group and $k \times (1-b)$ from the remaining samples. Each class-specific candidate group (for class $c$) is identified via agglomerative hierarchical clustering with Euclidean distances and Ward linkage of the $c$-labeled samples until a cluster of size $\geq k$ is obtained, from which $k \times b$ samples are drawn uniformly at random. The $k \times (1-b)$ samples are drawn uniformly at random from the remaining $c$-labeled samples. }
\label{fig:bias_induction_framework}
\end{figure}

Hierarchy bias generates a biased selection of samples for a given dataset, aiming to deviate from the original data distribution by skewing the representation of a group of samples that is deemed closer together in feature space than the remaining samples  (\figref{fig:bias_induction_framework}). The approach selects $k$ samples per class and controls group representation using bias ratio $b$ as follows. A class-specific group of at least $k$ closely related samples is first identified using agglomerative hierarchical clustering. To obtain the biased selection, $k \times b$ samples are chosen uniformly at random from the identified group and $k \times (1-b)$ samples are chosen uniformly at random from the remaining samples (\hyperref[sec:methods_hierarchy_bias]{Methods}).

\begin{figure}[tbh!]
\centering
\includegraphics[width=0.90\linewidth]{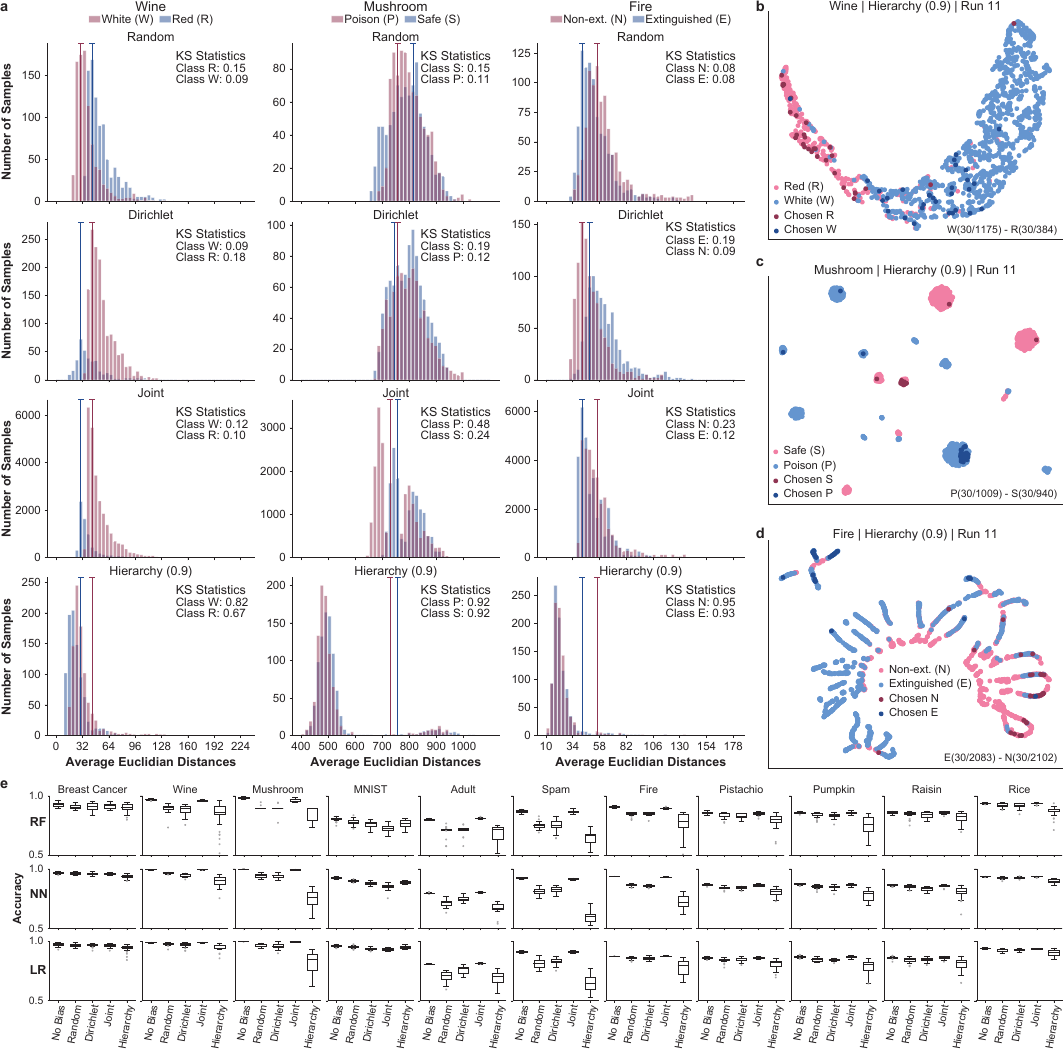}
\caption{\textbf{Bias induction impact on sample distances, latent space, and classifier performance.} 
\textbf{(a)} Class-specific distributions of per sample average Euclidean distances to all other samples, for the biased selection (histograms) and for all samples in the labeled train set (histogram peaks denoted by lines ending in a ``T'' shape), 
using three bias induction techniques (hierarchy with $b=0.9$, joint, and Dirichlet) and random subsampling on three datasets (wine, mushroom, and fire). 
Kolmogorov-Smirnov (KS) effect sizes quantify the distribution shift between the biased selection vs. all samples.
\textbf{(b-d)} Samples selected by hierarchy bias ($b=0.9$), highlighted on the respective latent UMAP space of the labeled train set for the wine, mushroom, and fire datasets (arbitrarily chosen run 11). \textbf{(e)} Accuracy of supervised RF, NN, and LR models on the test set after training on the original or biased labeled train set, over 30 distinct train runs. Box height delimits the interquartile range ($IQR = Q3-Q1$), with a line across the box denoting the median; whiskers indicate the largest and smallest values within $Q1-1.5 \times IQR$ and $Q3+1.5 \times IQR$, with points beyond the range as outliers. } 
\label{fig:bias_induction_results}
\end{figure}

To evaluate bias induction, we assessed the ability to generate a distribution shift between the biased selection and the original data, as well as the impact of the induced shift on ML model prediction performance. We compared hierarchy bias with $b=0.9$ to random subsampling and two alternative bias induction techniques: joint bias~\cite{Huang2006} and Dirichlet bias~\cite{Liu2014}. Hierarchy bias and random subsampling were set to select 30 samples per class, whereas Dirichlet targeted 60 and 300 samples in total respectively for binary and multiclass labeled datasets. Note that Dirichlet and joint bias do not take class labels into account when performing their selection, and joint bias does not allow control over the selected number of samples. 

\paragraph{Effect on data distribution.}
We first assessed the effect of bias induction on the distribution of distances between samples. The underlying idea is that a biased selection would exclude portions of the original data that deviate from the rest of the samples to some extent, thus making inter-sample distances closer on average. For each dataset, we obtained class-specific distributions of the per sample average Euclidean distance to all other samples. We further quantified the deviation between the class-specific distance distributions obtained for the biased selection and the original labeled set using Kolmogorov-Smirnov (KS) tests.
Hierarchy bias ($b=0.9$) induced the most significant shift in the distance distributions for all 11 datasets (KS effect sizes $> 0.65$, $p$-values $<0.05$; Fig. \ref{fig:bias_induction_results}a and Supplementary Fig. S1-S2), and primarily towards smaller average inter-sample distances, in line with the selection of close samples that hierarchy bias is designed to produce. Random selection resulted in the most similar distance distributions to the original data, with the smallest KS effect for 8 datasets. Dirichlet and joint bias led to modest shifts than hierarchy bias, with joint bias generally showing larger KS effects than Dirichlet (9 of 11 datasets). 
We also examined the samples selected from each labeled train set in the feature space, reduced to 2 dimensions (2D) using Uniform Manifold Approximation and Projection (UMAP) for an example run 11. Hierarchy bias selected samples from specific clusters or regions of the feature space. This was apparent across datasets (Supplementary Fig. S3), for instance hierarchy bias ignored samples in the top right area of the 2D space for the wine dataset (\figref{fig:bias_induction_results}b), selected from specific clusters of the mushroom dataset (\figref{fig:bias_induction_results}c), and focused on the top left and bottom right areas of the 2D space for the fire dataset (\figref{fig:bias_induction_results}d). In contrast, samples selected by random selection, as well as by the Dirichlet and joint biases, were spread throughout the 2D space and thus more representative of the original labeled train set for all datasets (Supplementary Fig. S4-S6). For random sampling, this was expected, given that no particular bias was introduced. For joint bias the result was also unsurprising, seeing that it selected the largest proportions of samples across datasets and thus captured most of the data (overall mean average 63\%, minimum 44\%, and maximum 80\%; for hierarchy bias: 17\%, 0.4\%, and 67\%; 
Supplementary Table S1). 

\paragraph{Impact on prediction performance.} We evaluated the impact of bias induction on the classification accuracy of supervised ML models for the 11 datasets across 30 runs. Per run, we trained 2-hidden layer neural network (NN), random forest (RF), and logistic regression (LR) models using the original labeled train set (No Bias) or a selection of its samples. The latter was obtained either by random subsampling or using Dirichlet, joint, or hierarchy bias induction. All models were evaluated on the original test set. The induced bias led to a decrease in accuracy with every technique except joint bias (\figref{fig:bias_induction_results}e), which as previously mentioned selected most of the original samples and thus did not induce particularly strong bias. Hierarchy bias caused the largest decrease in accuracy for all datasets except MNIST, where the most impact was seen with joint bias (\figref{fig:bias_induction_results}e). Note that the preset targets on the number of samples to select for hierarchy bias, Dirichlet bias, and random selection led these methods to select 64-70\% of the MNIST samples per class compared to 46-60\% with joint bias. This larger coverage of the original data likely influenced the ability of hierarchy and Dirichlet to produce a more effective biased selection for MNIST. 
Overall, hierarchy bias consistently selected samples in close proximity, leading to a significant shift in inter-sample distances and a bias towards class-specific parts of the original distribution. 
This caused a marked decrease in prediction accuracy of supervised ML models relative to other bias induction techniques. 

\subsection*{Diverse class-aware self-training (DCAST) for selection bias mitigation}

\begin{figure}[h!]
\centering
\includegraphics[width=\linewidth]{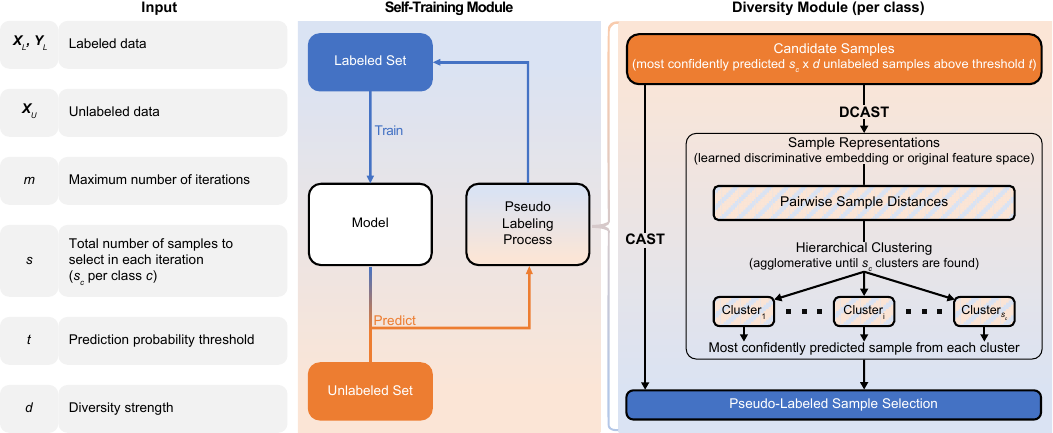}
\caption{\textbf{Diverse Class-Aware Self-Training (DCAST) framework.} 
(Left) Input to DCAST. 
Labeled data $\boldsymbol{X_L}$ (with labels $\boldsymbol{Y_L}$) and unlabeled data $\boldsymbol{X_U}$, maximum number of iterations $m$, number of pseudo-labeled samples $s$ to select per iteration, confidence or prediction probability threshold $t \in [0,1]$, and integer diversity strength parameter $d \geq 1$. 
(Middle) Self-training module. At each iteration, a model trained with labeled samples is used to predict pseudo-labels for unlabeled samples, from which a subset is newly selected and added to the labeled set for the next iteration. 
(Right) Diversity module. 
Selects the subset of $s_c = s \times class\_ratio(c)$ confidently predicted and diverse pseudo-labeled samples per class $c$, as follows: (i) select the top $s_c \times d$ samples from the unlabeled set with confidence or prediction probability larger than $t$ (or $1.2/C$, whichever is largest); and (ii) reduce this $s_c \times d$ selection to a set of $s_c$ diverse samples by identifying $s_c$ clusters using hierarchical clustering (agglomerative single-linkage) and selecting the most confidently predicted sample from each cluster. Note that $class\_ratio$ can otherwise be fixed to be equal across classes. Distance between samples is based on either learned discriminative embeddings, relating samples with respect to prediction output, or alternatively an unsupervised embedding or the original feature space. When $d=1$, DCAST becomes CAST, without the diversity strategy.}
\label{fig:bias_mitigation_framework}
\end{figure}

The proposed (D)CAST semi-supervised learning strategies (\figref{fig:bias_mitigation_framework}) aim to mitigate selection bias by leveraging insight from unlabeled data about the underlying distribution of the population. Both rely on self-training to gradually incorporate unlabeled data: at each training iteration, the learnt model is used to predict pseudo-labels for all unlabeled samples, from which a subset of $s$ samples ($s_c$ per class) is selected to be included in the labeled set for the next iteration. To address class-related bias, sample selection is done separately per class as follows. First, a set of $s \times d$ candidates is selected as the most confidently predicted samples with prediction probability above a threshold $t$, where $s$ and $d$ denote the number of samples to select and diversity strength. For CAST ($d=1$), this directly results in the final set of $s$ pseudo-labeled samples to add for the next iteration. The DCAST selection ($d>1$) extends upon CAST to mitigate confidence-related bias through sample diversity, reducing the set of $s \times d$ candidates to a final set of $s$ diverse pseudo-labeled samples. Capturing diverse sample groups is achieved via hierarchical clustering of the candidate samples into $s$ clusters ($s_c$ per class), followed by selection of diverse samples comprising the most confidently predicted sample per cluster. To ensure (D)CAST remains model-agnostic, sample distances for clustering can be based on discriminative embeddings learnt by the model or the original feature space.

\begin{figure}[ht!]
\centering
\includegraphics[width=\linewidth]{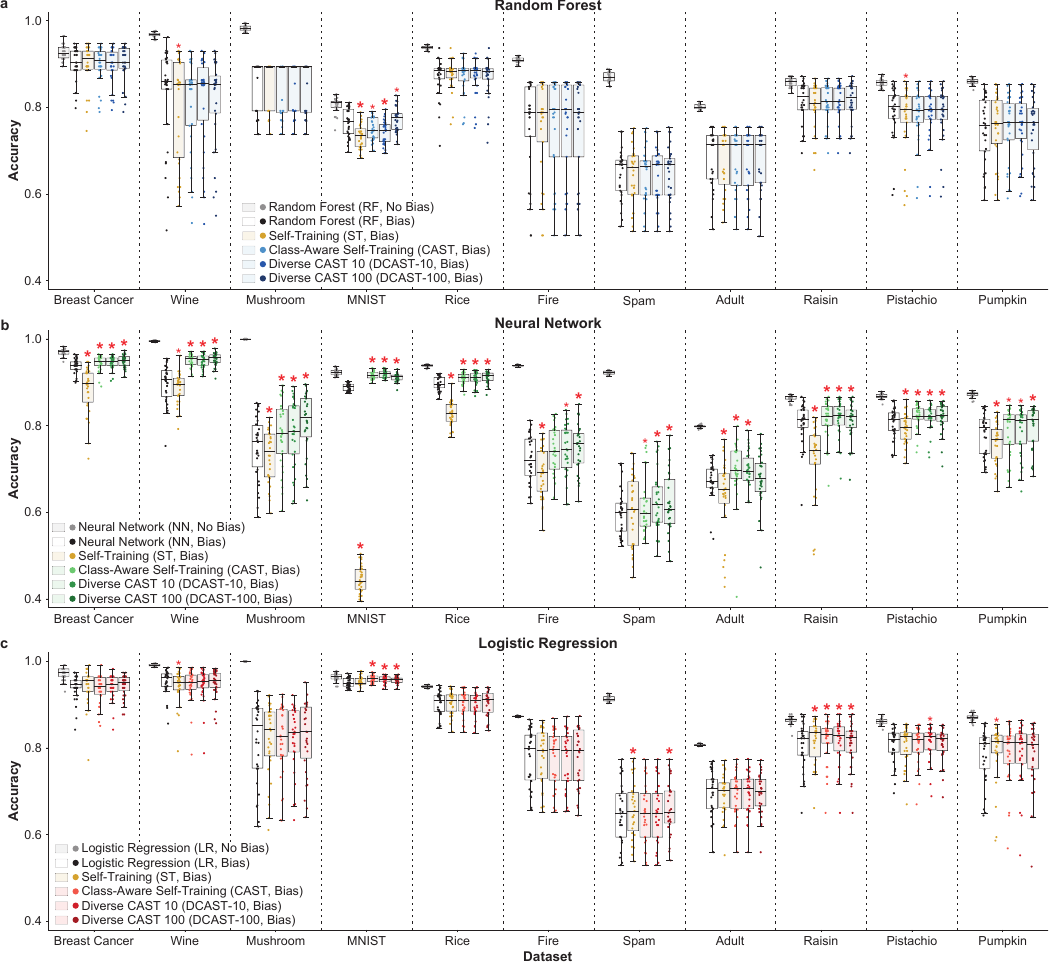}
\caption{\textbf{Bias mitigation by semi-supervised (D)CAST in the presence of hierarchy bias (ratio $b=0.9$).}
Accuracy of supervised and semi-supervised learning methods with \textbf{(a)} RF, \textbf{(b)} NN, and \textbf{(c)} LR models across 11 datasets. Results for 30 runs: each training on a different split of the train set into labeled and unlabeled sets, all evaluated on the same original test set. Models included (top to bottom): supervised RF/NN/LR models trained on the original (No Bias) or biased (Bias) labeled set; and semi-supervised RF/NN/LR models, using conventional self-training (ST) on the biased labeled train set plus the unlabeled test set, or (D)CAST on the biased labeled train set plus the unlabeled train set. 
Red asterisks (*) denote statistically significant changes in accuracy over 30 runs for each semi-supervised approach compared to supervised learning on the biased labeled set, using one-sided Wilcoxon signed-rank tests (larger asterisks indicate $p < 0.01$ and smaller asterisks $0.01 < p < 0.05$).}
\label{fig:semisupervised_bias_mitigation_results}
\end{figure}

\subsection*{Diversity and class-awareness in (D)CAST improve bias mitigation via self-training} 
To evaluate (D)CAST bias mitigation, we first assessed its test prediction accuracy against supervised learning and conventional self-training (ST)~\cite{McLachlan1975} on the biased labeled train set, with additional unlabeled samples for self-training strategies. Training and evaluation were performed for 11 datasets over 30 runs as previously described, using RF, NN, and LR models. We induced hierarchy bias with ratio $b=0.9$, as this type of selection bias showed the most impact on supervised models compared to Dirichlet and joint bias (\figref{fig:bias_induction_results}e). The (D)CAST method was assessed without diversity (CAST, $d=1$) or with diversities $d=\{10,100\}$ (CAST-10, DCAST-100), and was set to include $s=3\times$(number of classes) pseudo-labeled samples per iteration, for at most $m=100$ iterations, using prediction threshold $t=0.9$ (or the 85th or 93rd percentile in the case of RF models). 
Conventional ST selected the $3\times$(number of classes) most confidently predicted samples per iteration (\hyperref[sec:methods_experimentaldetails]{Methods, Bias mitigation strategies}).
Concerning the mitigation of hierarchy bias with ratio $b=0.9$, with NN models the semi-supervised (D)CAST strategies significantly improved generalizability over supervised learning across all 11 datasets ($p<0.05$ with one-sided Wilcoxon signed-rank tests, \figref{fig:semisupervised_bias_mitigation_results}b). Specifically, class-awareness with moderate diversity (DCAST-10) was significantly better than supervised learning on the 11 datasets, whereas class-awareness alone (CAST) or coupled with stronger diversity (DCAST-100) both improved on 10 datasets and remained comparable respectively on the fire and adult datasets. By contrast, conventional ST was significantly worse than supervised learning on 10 datasets with NN models. 
Using RF and LR models, mitigation of hierarchy bias with ratio $b=0.9$ was more modest. Semi-supervised (D)CAST and ST performed comparably to supervised learning on most datasets (8 with RF and 7 with LR models; \figref{fig:semisupervised_bias_mitigation_results}a,c), possibly due to the use of regularization, which could hamper model adaptation. 
We thus saw occasional statistically significant changes and smaller effect sizes with RF and LR models. Notably, the higher diversity strategy DCAST-100 led to the only significant improvement of semi-supervised over supervised learning using RF models, on the MNIST dataset (\figref{fig:semisupervised_bias_mitigation_results}a). Also with RF models, CAST and DCAST-10 decreased accuracy on MNIST, while ST decreased accuracy on 3 datasets (wine, MNIST, and pistachio; \figref{fig:semisupervised_bias_mitigation_results}a). 
With LR models, (D)CAST strategies improved over supervised learning on 4 datasets (MNIST, spam, raisin, and pistachio), whereas ST improved on 3 datasets (spam, raisin, and pumpkin) but also caused a decrease on the wine dataset (\figref{fig:semisupervised_bias_mitigation_results}c). 

Experiments with alternative bias induction techniques revealed similar findings, where (D)CAST bias mitigation consistently outperformed ST across datasets under random subsampling (Supplementary Fig. S7), and under induced Dirichlet or joint bias (Supplementary Figs. S8-S9). Again, we saw the largest performance differences with NN models, coinciding with the most improvement of (D)CAST and weakest results of ST over supervised learning. 

In summary, (D)CAST effectively mitigated selection bias induced by different techniques when paired with non-regularized NN models, and was not outperformed by supervised learning or conventional ST with regularized RF and LR models. In contrast, conventional ST struggled to recover from the bias with all three types of models, especially NNs. These results suggest that the class-awareness and diversity features introduced to the pseudo-labeling procedure in (D)CAST provide a promising semi-supervised learning strategy to mitigate selection bias.

\begin{figure}[ht!]
\centering
\includegraphics[width=\linewidth]{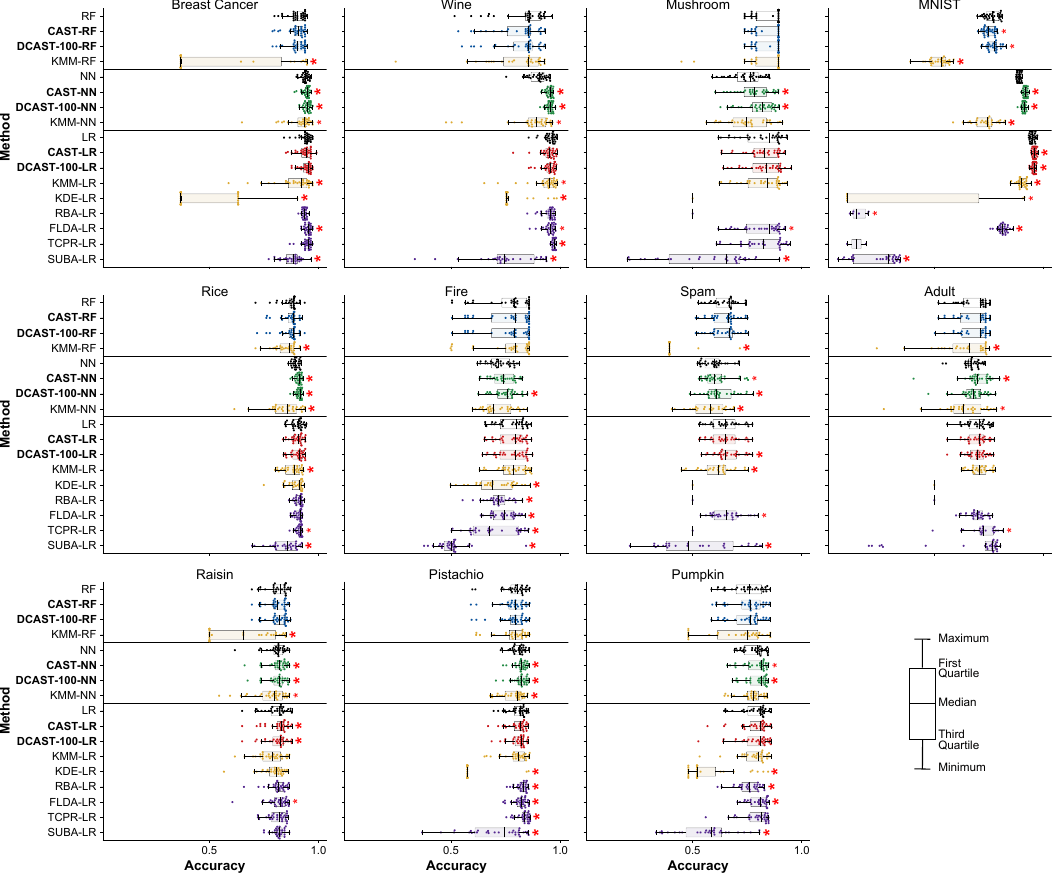}
\caption{\textbf{Bias mitigation by (D)CAST or domain adaptation beyond semi-supervised learning under hierarchy bias ($b=0.9$).}
Accuracy of semi-supervised (D)CAST strategies against alternative bias mitigation techniques with 3 different types of ML models for 11 datasets over 30 runs. Per run, each model was trained using a different labeled train set with induced hierarchy bias. We included a supervised learning model as baseline per ML model type (RF, NN, LR), together with bias mitigation models incorporating additional unlabeled samples from either the unlabeled train set ((D)CAST) or the unlabeled test set (remaining methods). All models were evaluated on the same original test set. 
Bias mitigation methods per category: semi-supervised (CAST and DCAST-100); importance weighting (KMM, KDE); minimax estimation (RBA, TCPR); and subspace alignment (FLDA, SUBA). The (D)CAST and KMM methods were coupled with RF, NN, and LR models, whereas the remaining methods used LR models only.
For clarity, horizontal lines group bias mitigation strategies by model type. The ``x'' symbol indicates model training was unsuccessful across all 30 runs.  } 
\label{fig:sota_bias_mitigation_results}
\end{figure}

\subsection*{Semi-supervised (D)CAST bias mitigation is superior to competing domain adaptation}
We also evaluated (D)CAST against bias mitigation techniques beyond semi-supervised learning. This included importance weighting methods KMM~\cite{Huang2006} and KDE~\cite{Shimodaira2000}, minimax approaches RBA~\cite{Liu2014} and TCPR~\cite{Kouw2021tcpr}, and subspace alignment methods FLDA~\cite{Kouw2016} and SUBA~\cite{Fernando2013}. All methods were trained on the biased labeled train set and evaluated on the original test set, with (D)CAST further incorporating samples from the unlabeled train set and the remaining methods using unlabeled test samples during training. The (D)CAST and KMM approaches were coupled with RF, NN, and LR models, while the remaining methods used LR only as per the original work.

Similar to our previous findings, CAST and DCAST-100 were the most robust bias mitigation methods. Overall, these strategies preserved or significantly improved over the supervised learning performance across the 3 model types and  11 datasets, with the exception of CAST showing a decrease in accuracy for MNIST when used with RF models. (\figref{fig:semisupervised_bias_mitigation_results}-\ref{fig:sota_bias_mitigation_results}).
In contrast, KMM led to significant decreases in accuracy for 8 datasets with NN models, as well as for 5 and 6 datasets respectively with LR and RF models.
As for the remaining bias mitigation methods using only LR models, KDE resulted in significant decreases in performance for all except the rice dataset. Apart from an improvement with RBA for the pistachio dataset, the RBA and SUBA methods degraded performance significantly for 6 and 9 datasets, respectively.  
The best competing methods were FLDA and TCPR, which showed significant improvements respectively for 5 and 4 datasets (FLDA: breast cancer, spam, raisin, pistachio, and pumpkin; TCPR: wine, rice, adult, and pistachio). The FLDA approach also led to significant decreases for 4 datasets (wine, mushroom, MNIST, and fire), while TCPR caused a significant decrease for the fire dataset. Concerning the MNIST dataset, TCPR failed to build models for most runs and caused a clear performance drop for the few remaining ones, resulting in insufficient power to determine statistical significance. Overall, CAST and DCAST-100 demonstrated consistent ability to match or outperform supervised learning in the presence of hierarchy bias compared to other bias mitigation methods. The gap was most evident on the multi-class classification problem (MNIST), where the other methods resulted in drastic decreases in performance.

\section*{Conclusion}
We put forth two contributions to improve the learning of prediction models in the presence of selection bias. First, a bias induction approach termed hierarchy bias to enable the evaluation of complex multivariate bias effects on the generalizability of prediction models. Second, a model-agnostic semi-supervised learning framework named (D)CAST that exploits unlabeled data in a class-aware manner and promotes sample diversity to mitigate selection bias. 

Hierarchy bias uses clustering to isolate one distinct group of samples per class and then skews the representation of such group during sample selection to induce class-specific multivariate bias, allowing control over the level of bias through a bias ratio parameter. Induced hierarchy bias showed a stronger impact on the distribution of inter-sample distances and proved more challenging for prediction models to overcome, compared to joint and Dirichlet bias. 

The (D)CAST model learning strategy progressively incorporates unlabeled samples using self-training, which is further made class-aware in CAST by pseudo-labeling confidently predicted unlabeled samples over a given threshold per class. Its extended variant, DCAST, seeks to counter confidence-associated bias with sample diversity by clustering and selecting pseudo-labeled samples from distinct groups, using distances based on either the discriminative embeddings provided by the underlying model or the original feature representation. 

Both class-awareness and diversity proved effective, leading to significant improvements in the bias mitigation ability of (D)CAST over conventional self-training across datasets and bias induction techniques. Models trained by (D)CAST also outperformed other models built using six alternative domain adaptation methods, comprising different importance weighting, minimax estimation, and subspace alignment approaches. 

Diversity strength was shown to influence the extent of (D)CAST bias mitigation, where a larger value resulted in improved robustness to selection bias. 
More generally, we recommend setting the diversity strength parameter such that the number of candidate samples considered for selection at each iteration is significantly larger than the number of samples to select. 
We further suggest choosing a number of samples to select per iteration comfortably below the size of the training set to promote a gradual adaptation of the model, but not too small so that the added samples can have an impact: a possible choice could be the closest even number to $\lfloor \sqrt{N} \rfloor$, with $N$ denoting the size of the training set. 
The confidence threshold can be adjusted according to the distribution of prediction probabilities of the model to allow (D)CAST 
to consider at least as many samples as the number to add at each iteration. 

We demonstrated that (D)CAST is model-agnostic through application with random forests (RF), neural networks (NN), and logistic regression (LR) models. The success of bias mitigation differed across architectures, with the most benefit achieved using NN models. We hypothesized that the use of regularization could also have played a role, by restricting model adaptation and thus limiting the contribution of unlabeled samples in the RF and LR models. Further investigation would be needed to obtain conclusive evidence. 

Overall, our results present  (D)CAST and hierarchy bias as promising strategies to improve the learning and evaluation of machine learning models in the presence of selection bias, as an essential step in striving towards fairness in machine learning.

\section*{Methods}
\label{sec:methods}

\subsection*{Hierarchy bias induction and (D)CAST bias mitigation}
\label{sec:methods_notations}
\paragraph{Notation.} We denote the input data (sample $\times$ feature) matrix as $\boldsymbol{X} \in \mathbb{R}^{N \times {F}}$, the input label matrix as $\boldsymbol{Y}\in \{0, 1\}^{N \times {C}}$, and output prediction probability matrix as $\boldsymbol{\bar{Y}} \in \mathbb{R}^{N \times {C}}$, where $N$ is the number of samples, $F$ is the number of features, and $C$ is the number of classes. Following this notation, $\boldsymbol{x}_n \in \mathbb{R}^{1 \times {F}}$ is the feature vector of sample $n \in \{1,2,...,N-1,N\}$, $y_n^c$ is the binary label of sample $n$ for class $c \in \{1,2,...,C-1,C\}$ (1 if assigned, 0 otherwise), and $\bar{y}_n^c$ is the prediction probability of sample $n$ being of class $c$ where $\sum_{c=1}^{C}{y_n^c} = 1$ and $\sum_{c=1}^{C}{\bar{y}_n^c} = 1$. 

\subsubsection*{Hierarchy bias}
\label{sec:methods_hierarchy_bias}
Hierarchy bias induction generates a biased selection of samples from a given dataset in a class-aware and multivariate manner. The idea is that the samples belonging to each class in the dataset can be seen as originating from a mixture of multivariate distributions. Based on this, the goal is to identify one of the mixtures and then make a skewed selection of samples by controlling the representation of the target mixture over the remaining samples. 
Hierarchy bias induction takes as input a data matrix $\boldsymbol{X}$, a label matrix $\boldsymbol{Y}$, a parameter $k$ denoting the number of samples to select per class, and a bias parameter $b \in [0,1]$ denoting the ratio of samples that should be selected from the identified mixture (\algoref{alg:hierarchy_bias}). The output is a biased selection of samples, generated as follows. Agglomerative hierarchical clustering is first applied to identify a mixture of interest per class $c$, corresponding to a cluster of at least $k$ samples. We perform the clustering for class $c$ using all samples from matrix $\boldsymbol{X}$ labeled with class $c$, with Euclidean inter-sample distances on the original feature vectors and Ward linkage between clusters (Alg. \ref{alg:hierarchy_bias}, lines 4-5). Once the cluster is identified, the final biased selection is obtained by choosing $k \times b$ samples uniformly at random from the cluster and choosing another $k - k \times b$ samples uniformly at random from the remaining samples not in the cluster  (Alg. \ref{alg:hierarchy_bias}, lines 6-8).

\begin{algorithm}
\caption{Hierarchy Bias}\label{alg:hierarchy_bias}
\begin{algorithmic}[1]
\Require $\boldsymbol{X}$, $\boldsymbol{Y}$, $k$, $b$.
\Ensure $Selection \gets \emptyset$
\State $k_{cluster} \gets k \times b$
\State $k_{rest} \gets k-k \times b$
\For{\texttt{each class} $c \in C$}
    \State Apply agglomerative clustering with Euclidean distance and Ward linkage to $\boldsymbol{X}_{S_c}$, $S_c=\{ n : n \in y_n^c==1\}$.
    \State $Cluster \gets$ Set of samples from the first cluster that reaches a number of samples $\geq k$. 
    \State $S_{cluster} \gets$ Select set of $k_{cluster}$ samples uniformly at random from $Cluster$.
    \State $S_{rest} \gets$ Select set of $k_{rest}$ samples uniformly at random from the remaining samples (not in $Cluster$).
    \State $Selection \cup 
    S_{cluster} \cup S_{rest}$
\EndFor
\State \Return $Selection$
\end{algorithmic}
\end{algorithm}

\subsubsection*{(D)CAST - Diverse Class-Aware Self-Training}
\label{sec:methods_dcast}
The proposed semi-supervised model learning framework, Diverse Class-Aware Self-Training (DCAST), leverages unlabeled data to gain insight into the underlying distribution of the population that may not be well represented by the labeled data. It does this using self-training (ST), and actively addresses selection bias by preserving class ratios or balance (CAST), and optionally also incorporating sample diversity into the pseudo-labeling process to counter biases present in the data or introduced during training (DCAST). 

More formally, the (D)CAST method takes as input the labeled data $\{\boldsymbol{X_L},\boldsymbol{Y_L}\}$ and unlabeled data $\boldsymbol{X_U}$ to learn from, validation data $\{\boldsymbol{X_V},\boldsymbol{Y_V}\}$ for early stopping, and the following four additional parameters: maximum number of iterations $m$, number of pseudo-labeled samples $s$ to select per iteration, confidence or prediction probability threshold $t \in [0,1]$, and integer diversity parameter $d \geq 1$. 
Model learning in (D)CAST is then performed by self-training as follows. 
At iteration $i$, model $M^{(i)}$ is trained on the labeled data $\{\boldsymbol{X_{L^{(i)}}},\boldsymbol{Y_{L^{(i)}}}\}$, and used to make predictions $\boldsymbol{\bar{Y}_{U^{(i)}}}$ for all samples in the unlabeled set $U^{(i)}$ (and matrix ${\boldsymbol{X_{U^{(i)}}}}$). 
As with regular self-training, a pseudo-labeling procedure then selects a subset of the unlabeled samples, $S^{(i)} \subseteq U^{(i)}$, to be incorporated into model learning (\figref{fig:bias_mitigation_framework}). The selected samples $S^{(i)}$ are pseudo-labeled and included in the set of labeled samples for training in the subsequent iteration, $L^{(i+1)} = L^{(i)} \cup S^{(i)}$, as well as removed from the unlabeled set $U^{(i+1)} = U^{(i)} \setminus S^{(i)}$. Matrices ${\boldsymbol{X_{L^{(i+1)}}}}$,
${\boldsymbol{Y_{L^{(i+1)}}}}$, and ${\boldsymbol{X_{U^{(i+1)}}}}$ are also updated for the next iteration accordingly. 

\paragraph{Pseudo-labeling in (D)CAST: class-aware with and without diversity.} 
\label{sec:methods_dcast_labeling}
The (D)CAST-specific pseudo-labeling is accomplished by the Diversity Module (\figref{fig:bias_mitigation_framework}). The core CAST strategy addresses class-specific bias by performing the pseudo-labeling separately per class, offering to either preserve the class ratios found in the original labeled set or select an equal number of samples per class at each iteration.
Its extension, DCAST, aims for further bias mitigation by promoting sample diversity. In conventional self-training, the pseudo-labeling procedure tends to confirm and follow biases potentially present in the labeled set: either by selecting unlabeled samples similar to the original labeled samples (in feature space) or by selecting unlabeled samples whose prediction the model is most confident about.
In contrast, (D)CAST seeks to mitigate this behavior and work against the strengthening of existing bias during training. To achieve this, (D)CAST selects and pseudo-labels samples that are diverse amongst each other and also more dissimilar to the possibly biased labeled samples. The (D)CAST pseudo-labeling (Alg. \ref{alg:DCAST}) comprises the following steps per training iteration:

\textbf{Step 1. (D)CAST - Select candidate samples for pseudo-labeling based on model confidence.} The goal of Step 1 is to select a set of candidate unlabeled samples for pseudo-labeling and inclusion in model training. This corresponds to the $s \times class\_ratio(c) \times d$ most confidently predicted unlabeled samples per class $c$, with corresponding probabilities in $\boldsymbol{\bar{Y}_{U^{(i)}}}$ larger than a user-defined threshold $t$ (or a baseline threshold $r = 1.2/C$, whichever is largest) (Alg. \ref{alg:DCAST}, lines 9-11). For CAST, with $d=1$ and thus no diversity strategy, this selection automatically leads to the final set of $s$ pseudo-labeled samples ($s_c = s \times class\_ratio(c)$ per class) to incorporate during learning in the subsequent iteration. For DCAST, with $d>1$ (Alg. \ref{alg:DCAST}, lines 13-15), the selected set of $s \times d$ samples ($s_c \times d$ per class) represents a larger pool of candidates to consider and narrow down further to obtain the final selected set of $s$ samples ($s_c$ per class) using the diversity strategy. Our recommendation for DCAST is to set the confidence threshold $t$ and diversity parameter $d$ not too strictly, so as to allow for a sufficient number (and diversity) of candidate samples.

\textbf{Step 2. DCAST - Diversity: Create representations of candidate samples for distance calculation.} From the set of $s \times d$ candidate samples selected in Step 1, DCAST aims to extract the subset of $s$ diverse samples. Diversity is assessed based on pairwise sample distances, calculated using a specific sample vector representation or embedding (denoted for all candidate samples as matrix $\boldsymbol{E^{(i)}}\in \mathbb{R}^{(s \times d) \times {v}}$, where $v$ is the embedding vector size). Preferably, DCAST uses discriminative embeddings based on the learnt model $M^{(i)}$, where two types are currently supported. For a random forest, each sample representation corresponds to a one-hot encoded vector of the prediction of that sample across all the leaves of the decision trees in the forest; for a neural network, the sample representation corresponds to the embedding based on the hidden layer closest to the output layer. For models without discriminative embeddings, such as SVM or LR, DCAST uses the original feature vector representation.

\textbf{Step 3. DCAST - Diversity: Calculate pairwise distances between candidate samples.} To assess diversity, we use distances between samples: the larger the distances amongst samples in a given set, the more diverse the set will be considered. Distances are calculated by DCAST based on sample embeddings or original feature vector representations (Alg. \ref{alg:DCAST}, line 13). With discriminative embeddings, DCAST calculates normalized distances as ${1-(E{\cdot}E^T)/\max(E{\cdot}E^T)}$, given an embedding matrix $E\in \mathbb{R}^{(s \times d) \times v}$. Specifically, for a random forest model, these distances represent the normalized frequency of non co-occurrence of a pair of samples in the leaves of the decision trees. With original feature vectors, DCAST uses Euclidean distances between sample vectors instead.

\textbf{Step 4. DCAST - Diversity: Identify distinct clusters and select diverse samples to pseudo-label.} The distances calculated in Step 3 are used in Step 4 to select diverse samples, potentially capturing different aspects of the pool of candidates and its underlying distribution. To do this, DCAST first identifies $s$ (or $s_c$ per class) distinct groups of candidate samples using a clustering algorithm (Alg. \ref{alg:DCAST}, line 14). The current implementation relies on agglomerative hierarchical clustering with single linkage, however any other algorithm of choice could be employed. Given that clustering is designed to maximize inter-cluster distances, samples across the different clusters are likely to yield the largest distances and thus the most diversity under the employed clustering strategy.
Accordingly, DCAST selects a single sample per identified cluster to pseudo-label, namely the candidate sample with the highest confidence $\bar{y}^c_n$ value (sample $n$ and class $c$, Alg. \ref{alg:DCAST}, line 15).

\textbf{Step 5. (D)CAST - Pseudo-label selected samples.} 
At the end of each iteration, selected samples in the set $S_c$ are added to the labeled data matrices $\{\boldsymbol{X_L},\boldsymbol{Y_{L}}\}$ and removed from the unlabeled data matrix $\boldsymbol{X_U}$.  

\paragraph{Time Complexity of (D)CAST.} 
\label{sec:methods_dcast_timecomplexity}
To derive an upper bound for the worst-case time complexity of the (D)CAST algorithm, we assume the following time complexities for an input of $n$ samples defined over $v$ features: training a base prediction model is $O(T(n, v))$, making predictions using the trained model is $O(P(n, v))$, and calculating pairwise sample distances and applying hierarchical clustering is $O((n\times v)^2)$. 

At iteration $i$, the time complexity of (D)CAST is dominated by the following operations: retraining the model with $l+i \times s$ labeled samples in $O(T(l+i \times s, v))$ time (Alg. \ref{alg:DCAST}, line 4), making predictions for $l-i \times s$ unlabeled samples in $O(P(l-i \times s, v))$ time (Alg. \ref{alg:DCAST}, line 5), and applying hierarchical clustering with pairwise distances to at most $s \times d$ candidate unlabeled samples in $O((s\times d \times v)^2)$ time (Alg. \ref{alg:DCAST}, lines 11-12). Note that $l$ denotes the number of labeled samples in the input matrices $\{\boldsymbol{X_L},\boldsymbol{Y_{L}}\}$ at the start of the execution, and $i\times s$ denotes the number of samples that are pseudo-labeled up to iteration $i$ (thus also added and removed respectively from the labeled and unlabeled data). The maximum possible number of samples for prediction at any one iteration is equal to the number of unlabeled samples $u$ in the input matrix $\boldsymbol{X_U}$ before any pseudo-labeling has occurred, leading to the upper bound $O(P(u,v))$ on the prediction time per iteration. Similarly, $u$ is the maximum number of samples that can be added to the input labeled data (initially containing $l$ samples) over all iterations, which determines the upper bound $O(T(l+u,v))$ on the training time per iteration. Combining all together, each iteration takes  $O(T(l+u,v) + P(u,v)+ (s \times d \times v)^2)$ time, and therefore the upper bound on the worst-case time complexity of $m$ iterations is $O(m \times (T(l+u,v) + P(u,v)+ (s \times d \times v)^2))$.

\begin{algorithm}
\caption{(D)CAST - Diverse Class-Aware Self-Training}\label{alg:DCAST}
\begin{algorithmic}[1]
\Require $T$ (model type); $\boldsymbol{X_L}$, $\boldsymbol{Y_L}$ (labeled train data); $\boldsymbol{X_V}$, $\boldsymbol{Y_V}$ (labeled validation data); $\boldsymbol{X_U}$ (unlabeled data); $s$ (number of samples to select per iteration); $t$ (prediction probability threshold); $d$ (diversity strength); $m$ (maximum number of iterations).
\State $terminate \gets False$
\State $i \gets 0$
\While{$terminate$ is $False \lor i = m$}
    \State $M^{(i)} \gets $ train model instance of type $T$ with $\boldsymbol{X_L}$, $\boldsymbol{Y_L}$
    \State $\bar{Y} \gets$  predict class probability for samples in $\boldsymbol{X_U}$ using $M^{(i)}$
\For{\texttt{each class} $c \in C$}
    \State $s_c \gets s \times class\_ratio(c)$
    \State $t_c \gets \max(t, r)$
    \State $S_c \gets $ top $s_c \times d$ confidently predicted samples with 
    $max(\bar{y}_n^c) > t_c$ 
    \If {$d > 1$}
        \State $E \gets $ calculate pairwise distances for samples in $S_c$ 
        \State $Clusters \gets $ apply agglomerative clustering to obtain $s_c$ clusters using distances $E$ and single linkage
        \State $S_c \gets $ choose the sample with the highest prediction probability from each cluster in $Clusters$
    \EndIf
    \For{\texttt{each} selected sample $n \in S_c$}
        \State $\boldsymbol{X_L}.\textrm{add}(\boldsymbol{x_n})$, $\boldsymbol{Y_L}.\textrm{add}(\boldsymbol{y_n})$, $\boldsymbol{X_U}.\textrm{remove}(\boldsymbol{x_n})$
    \EndFor
\EndFor
    \LineComment{Stopping conditions: maximum number of iterations $m$ is reached OR all unlabeled samples have been incorporated OR validation accuracy did not improve for the last 5 iterations.}
    \If {
    ( $i == m$ ) $\lor$ \\
    ( $len(\boldsymbol{X_U} )==0$) $\lor$ \\ 
    ( $\exists z \in \{i-6, \ldots, i-1\}$  such that $Accuracy(M^{(i)}, \boldsymbol{X_V}, \boldsymbol{Y_V}) < Accuracy(M^{(z)}, \boldsymbol{X_V}, \boldsymbol{Y_V}$) )}
    
        \State \hspace{\algorithmicindent} $terminate \gets True$
        \State \hspace{\algorithmicindent} $M_{best} \gets argmax_{z=0, ..., i}(Accuracy(M^{(z)}, \boldsymbol{X_V} , \boldsymbol{Y_V}))$
    \EndIf
    \State $i \gets i+1$
\EndWhile
\State \Return $M_{best}$
\end{algorithmic}
\end{algorithm}

\subsection*{Data}
\label{sec:methods_data}
In addition to 8 datasets from the UCI Data Repository (breast cancer, adult, spam, wine, raisin, rice, mushroom, and MNIST; \href{https://archive.ics.uci.edu/}{https://archive.ics.uci.edu}), we also used 3 datasets from other sources, including the pistachio \cite{Ozkan2021_pistachio}, fire \cite{Koklu2021_fire}, and pumpkin \cite{Koklu2021_pumpkin} datasets (Supplementary Table S2). All datasets had binary class labels, except for MNIST with 10 different class labels. The breast cancer, wine, spam, rice, raisin, pistachio, pumpkin and MNIST datasets comprised between 7 to 64 continuous features. The fire and adult datasets included mixed types of features, of which 1 and 7 were respectively categorical features. The mushroom dataset only had categorical features. 
For the fire, adult, and mushroom datasets, all categorical features were one-hot encoded. 

\subsection*{Evaluation of bias induction and bias mitigation methods}
We performed experiments across 11 ML benchmark datasets with different characteristics to assess the effectiveness of (i) selection bias induction using the proposed hierarchy bias technique, and (ii) selection bias mitigation using the proposed (D)CAST strategies. Hierarchy bias was compared to other bias induction techniques concerning both the distribution shift produced by the data selection procedure and its effect on the performance of prediction models built using supervised learning. The (D)CAST semi-supervised bias mitigation strategies were evaluated against conventional semi-supervised self-training (ST), as well as a range of alternative domain adaptation methods, on their ability to build prediction models from biased data with better generalization than using supervised learning.  

\begin{figure}[ht!]
\centering
\includegraphics[width=\linewidth]{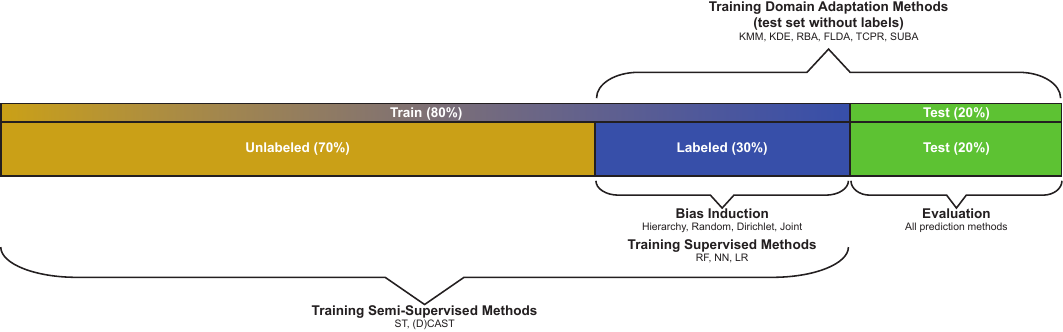}
\caption{\textbf{Data split for evaluation of bias induction and bias mitigation effects on prediction performance.}
Each dataset is randomly split into train (80\%) and test (20\%) sets, and 30 different train runs are created by splitting the samples in the train set randomly into labeled (30\%) and unlabeled (70\%) train sets. Bias induction is further applied to the labeled train sets to generate corresponding biased labeled train sets. Supervised learning is used to build models separately from the original labeled train set and from the biased labeled train set, which serve as baselines to assess the effects of bias induction and bias mitigationl on prediction performance. For bias mitigation, CAST and DCAST learn prediction models using both the unlabeled and labeled train sets, while domain adaptation methods learn from the labeled train set together with the test set (without labels). All models are evaluated on the labeled test set. }
\label{fig:data_split}
\end{figure}

\noindent\textbf{Data splits and bias induction.}
For each dataset, 20\% of the samples were uniformly selected at random, stratified by class, and reserved as test data to evaluate prediction models (\figref{fig:data_split}). The adult dataset already had its own separate test set, which we reserved.
Additionally, we created 30 distinct train runs per dataset, each by randomly splitting the remaining 80\% of the samples into two train sets, stratified by class: a labeled train set, containing 30\% of the samples, from which we also generated biased labeled sets by applying different bias induction techniques; and an unlabeled train set, comprising the remaining 70\% of the samples.
The original and biased labeled train sets were later used to build prediction models with supervised learning or bias mitigation strategies, while the unlabeled train set was used 
to learn prediction models with the semi-supervised bias mitigation strategies (D)CAST and conventional ST (other bias mitigation methods used test data without labels). 
When necessary for model training, a validation set was further extracted from each biased train set, given that unbiased labeled data would not be available for this purpose in a realistic setting.

\subsubsection*{Bias induction impact on data distribution}
Bias induction methods were first assessed on their ability to cause a distribution shift in the biased selection relative to the original labeled train set. Quantitatively, we analyzed the change in the distribution of inter-sample distances as follows. We first calculated class-specific distributions of the per sample 
average Euclidean distance to all other samples in either the biased selection or the original labeled train set. We then determined the class-specific distribution shifts between the biased selection and the original data using two-sample Kolmogorov-Smirnov (KS) statistical tests. We report KS effect sizes, as well as histograms of inter-sample distances for the biased selection distribution and histogram peaks for the original data distribution. 

Visually, we analyzed to what extent the biased selection was representative of the original labeled train set by inspecting 2D dimension reductions of the original data using the Uniform Manifold Approximation and Projection (UMAP) algorithm. We applied UMAP to the original labeled set with four different nearest neighbor parameter values (15, 50, 100, and 200) to obtain a reasonable representation of the sample space for each dataset. 

\subsubsection*{Bias induction and mitigation effects on prediction performance}
\label{sec:methods_experimentalsettings}
Furthermore, to evaluate bias induction and bias mitigation techniques, we investigated how prediction models trained on data affected or not by selection bias generalized to test data that was more representative of the original distribution. All models built using supervised learning or bias mitigation techniques were trained and evaluated as follows.  

\noindent\textbf{Training of models using supervised learning or bias mitigation.} To quantify the baseline prediction performance, without bias induction, we built models using supervised learning on the original labeled train set. To assess the effect of bias induction compared to the baseline, we built models using supervised learning on the biased labeled train set. Additionally, to assess the bias mitigation strategies and investigate if they could generalize better than supervised learning on the biased labeled train set, we used them to train models on the biased labeled train set together with unlabeled data (namely the unlabeled train set for semi-supervised (D)CAST and conventional ST, or the unlabeled test set for the remaining methods). The prediction models we trained using supervised learning or bias mitigation strategies were based on three different model types: L2-regularized random forests (RF, \cite{ho1995random}), 2 hidden-layered (input, 8-node, 12-node, output) neural
networks (NN), and L2-regularized logistic regression (LR) \cite{scikitlearn}. We used default parameter values (Supplementary Table S3), since fine-tuning with a biased validation set could further reinforce the bias. To account for variation introduced by randomness in the training procedures of the RF and NN models, we used different seeds to train 10 prediction models instead of one per run for any given combination of dataset, model type, bias induction technique, and model learning strategy. 

\noindent\textbf{Evaluation of models trained using supervised learning or bias mitigation.} The performance of all prediction models was evaluated on the test set. We focused on quantifying prediction accuracy rather than loss, since the loss could often be improved by increasing model confidence without a measurable improvement in accuracy, which is ultimately the goal of the models under study. We report the performance results as the median test accuracy of the 10 models using different seeds per run, with a total of 30 runs, for every combination of dataset, model type, bias induction technique, and model learning strategy. Some model learning strategies did not successfully build prediction models for all runs, which is necessarily reflected in the results and corresponding figures.

\subsection*{Experimental settings of bias induction and mitigation methods}
\subsubsection*{Bias induction and sample selection methods}
\label{sec:methods_otherinductionmethods}
We compared the proposed hierarchy bias induction method against the joint and Dirichlet bias induction techniques, as well as random subsampling. Hierarchy bias was used with a fixed target of $k=30$ samples to select per class, and a bias ratio of $b=0.9$ across experiments. Random subsampling consisted in selecting $k$ samples uniformly at random per class, where $k$ was similarly set to 30. 
Joint bias assigns a selection probability to each sample based on its proximity to the sample mean over the labeled train data, and then independently selects samples according to their selection probabilities~\cite{Huang2006}. Joint bias induction does not include any parameter to control the number of selected samples, and it was therefore used without a fixed target number of selected biased samples. Dirichlet bias selects a subset of samples without replacement, where the biased selection probability of each sample is determined based on a random likelihood function sampled from a Dirichlet distribution ~\cite{Liu2014}. 
This method does not consider class labels in its biased selection and was therefore set to select a total of $k \times |C|$ samples, with $|C|$ denoting the number of classes and $k=30$. Of note, hierarchy bias and random subsampling generate a biased selection that is balanced across classes, whereas joint and Dirichlet bias induction do not offer such guarantee. 

\subsubsection*{Bias mitigation strategies}
\label{sec:methods_experimentaldetails}
We assessed the proposed semi-supervised (D)CAST methods against competing bias mitigation techniques, including semi-supervised conventional self-training and alternative domain adaptation strategies. 

The semi-supervised methods, (D)CAST and conventional ST, learned models using the labeled and unlabeled train sets. Additionally, (D)CAST relied on early stopping based on validation performance to make training more efficient and robust. To be fair to other methods, (D)CAST used a portion of the labeled train set for validation rather than a separate validation set. 
We set the following parameter values for (D)CAST across experiments: maximum number of iterations $m = 100$, number of pseudo-labeled samples to include per iteration $s$ as $3 \times |C|$ (or 3 times the number of classes), and three different diversity strengths $d=\{1,10,100\}$. In addition, the confidence threshold $t$ used by (D)CAST to select candidate samples for pseudo-labeling was set to a prediction probability of $0.9$ for NN and LR models. Since RF models generally showed lower prediction probabilities, possibly due to regularization, we defined the threshold for binary RF classification models as the 93rd percentile of all prediction probabilities on unlabeled data. This threshold was not fully optimized, only considered sufficient to allow pseudo-labeling of some samples across all datasets with binary class labels. For MNIST, probabilities were even lower given the multiclass nature of the problem, thus we set the threshold of RF models as the 85th percentile instead.

Given that most semi-supervised learning approaches designed to mitigate sample selection bias are not model agnostic and do not have readily available implementations, we compared (D)CAST with the closely related conventional self-training (ST) methods.
We implemented and tested two variants of conventional ST, which pseudo-labeled either the $3 \times |C|$ samples with the highest prediction probabilities or all samples with prediction probabilities over 0.9. The former variant performed better and was thus selected.

We included domain adaptation methods beyond semi-supervised learning across three categories, using Python implementations available in the libTLDA Python library \cite{wouter_kouw_2018_1214315}: importance weighting approaches Kernel Mean Matching (KMM~\cite{Huang2006}) and Kernel Density Estimation (KDE~\cite{Shimodaira2000}), minimax estimation strategies Robust Bias-Aware classifier (RBA~\cite{Liu2014}) and Target Contrastive Pessimistic Risk (TCPR~\cite{Kouw2021tcpr}), and subspace alignment methods Feature-Level Domain Adaptation (FLDA~\cite{Kouw2016}) and Subspace Alignment classifier (SUBA~\cite{Fernando2013}). All of these methods were applied as originally proposed by their authors to learn models based on the labeled train set together with the test set without labels. In addition, all methods except KMM were used exclusively with L2-regularized LR models. The KMM importance weighting approach is ML model-agnostic, since it independently calculates a weight for each sample based exclusively on the train and test data, and was therefore applied with RF, NN, and LR models.

\section*{Data availability}
The data used in this article were obtained from publicly available sources, detailed in the Methods section. The raw data necessary to reproduce the experiments, along with the main experimental results for CAST and DCAST, are accessible via Figshare at \href{https://doi.org/10.6084/m9.figshare.27003601}{doi.org/10.6084/m9.figshare.27003601}.

\section*{Code availability}
An implementation of the hierarchy bias and the (D)CAST methods in Python has been made available under an open source license at \href{https://github.com/joanagoncalveslab/DCAST}{github.com/joanagoncalveslab/DCAST}.

\noindent\textbf{Acknowledgements}\\
The authors received funding from the US National Institutes of Health [U54EY032442, U54DK134302, U01DK133766, R01AG078803 to J.P.G.]. Authors are solely responsible for the research, the funders were not involved in the work. 
The authors further acknowledge the High-Performance Compute (HPC) cluster of the Department of Intelligent Systems at the Delft University of Technology.\\\\
\noindent\textbf{Author contributions}\\
Conceptualization, Y.I.T., and J.P.G.; Methodology, Y.I.T. and J.P.G.; Validation and Formal Analysis, Y.I.T.; Software, Y.I.T.; Investigation, Y.I.T. and J.P.G.; Writing – Original Draft, Y.I.T.; Writing – Review \& Editing, J.P.G.; Funding Acquisition and Supervision, J.P.G.  \\\\
\noindent\textbf{Competing interests}\\
The authors declare no competing interests. \\\\
\newpage

\end{document}


\flushbottom
\maketitle

\thispagestyle{empty}
\tableofcontents
\newpage

\section{Supplementary Figures}

\subsection{Bias induction to other datasets}
\begin{figure}[ht!]
\centering
\includegraphics[width=\linewidth]{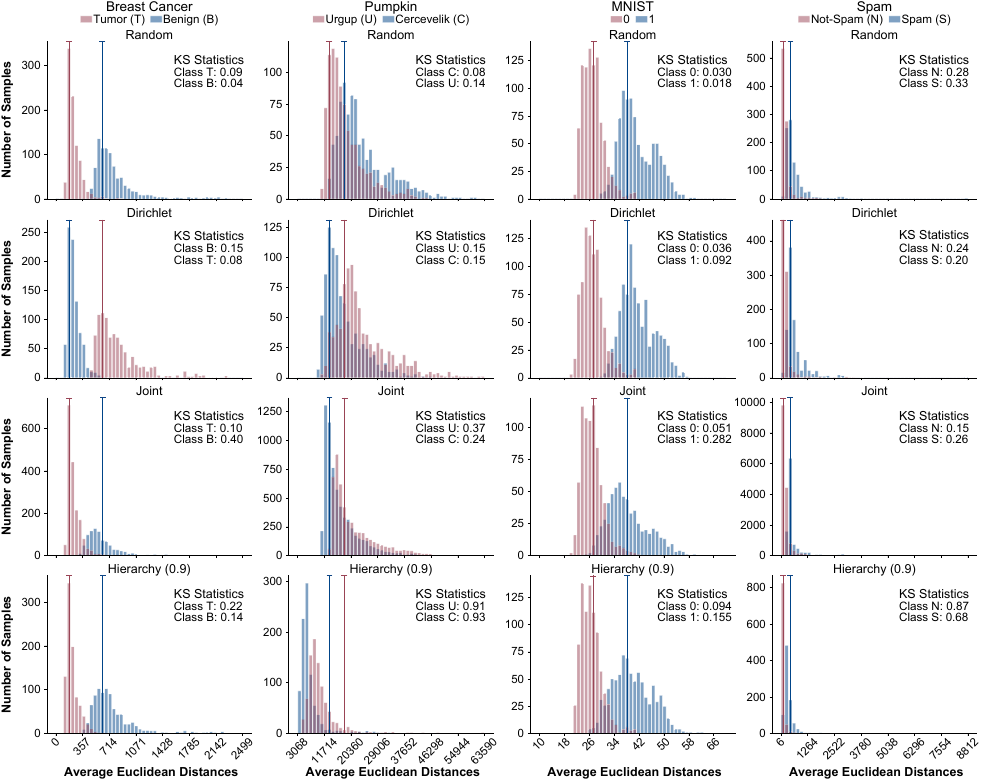}
\caption{\textbf{Bias induction impact on sample distances for the breast cancer, pumpkin, MNIST, and spam datasets.}
Class-specific distributions of per sample average Euclidean distances to all other samples, for the biased selection (histograms) and for all samples in the labeled train set (histogram peaks denoted by lines ending in a ``T'' shape), using three bias induction techniques (hierarchy with $b=0.9$, joint, and Dirichlet) and random subsampling on four datasets (breast cancer, pumpkin, MNIST, and spam). Kolmogorov-Smirnov (KS) effect sizes quantify the distribution shift between the biased selection vs. all samples distributions.
} 
\label{supfig:induction_histogram_breast_pumpkin_mnist_spam}
\end{figure}

\begin{figure}[th!]
\centering
\includegraphics[width=\linewidth]{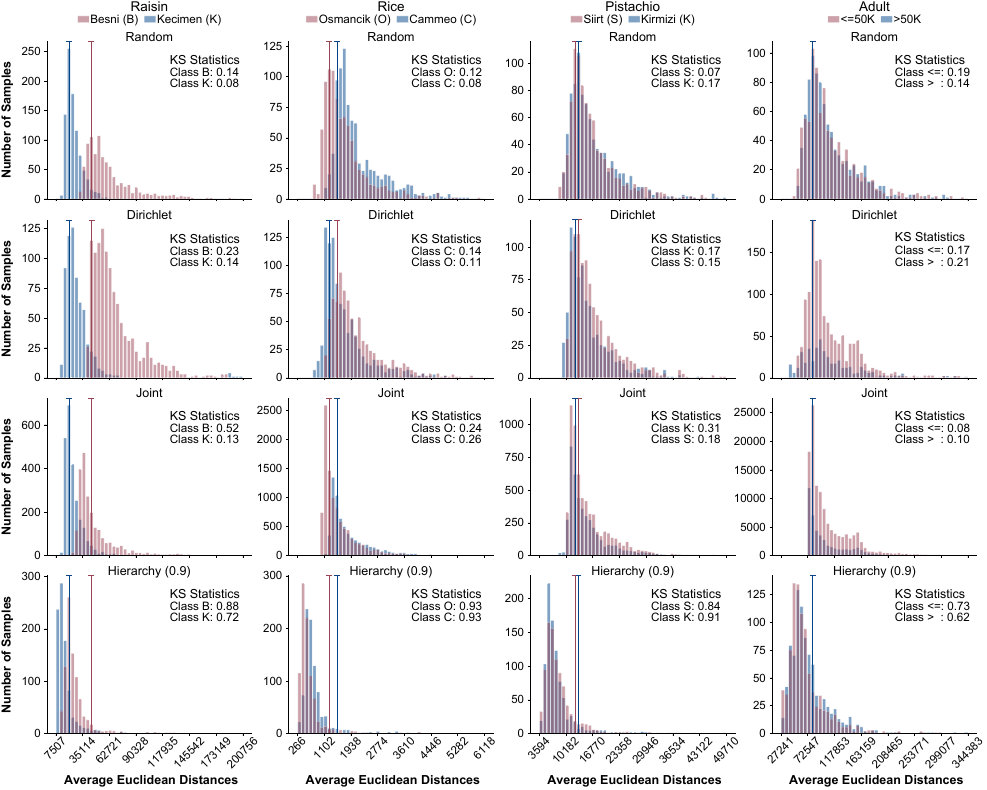}
\caption{\textbf{Bias induction impact on sample distances for the raisin, rice, pistachio, and adult datasets.}
Class-specific distributions of per sample average Euclidean distances to all other samples, for the biased selection (histograms) and for all samples in the labeled train set (histogram peaks denoted by lines ending in a ``T'' shape), using three bias induction techniques (hierarchy with $b=0.9$, joint, and Dirichlet) and random subsampling on four datasets (raisin, rice, pistachio, and adult). Kolmogorov-Smirnov (KS) effect sizes quantify the distribution shift between the biased selection vs. all samples distributions.
} 
\label{supfig:induction_histogram_raisin_rice_pistachio_adult}
\end{figure}

\clearpage

\begin{figure}[th!]
\centering
     \begin{subfigure}[b]{0.32\textwidth}
         \centering
         \includegraphics[width=\textwidth]{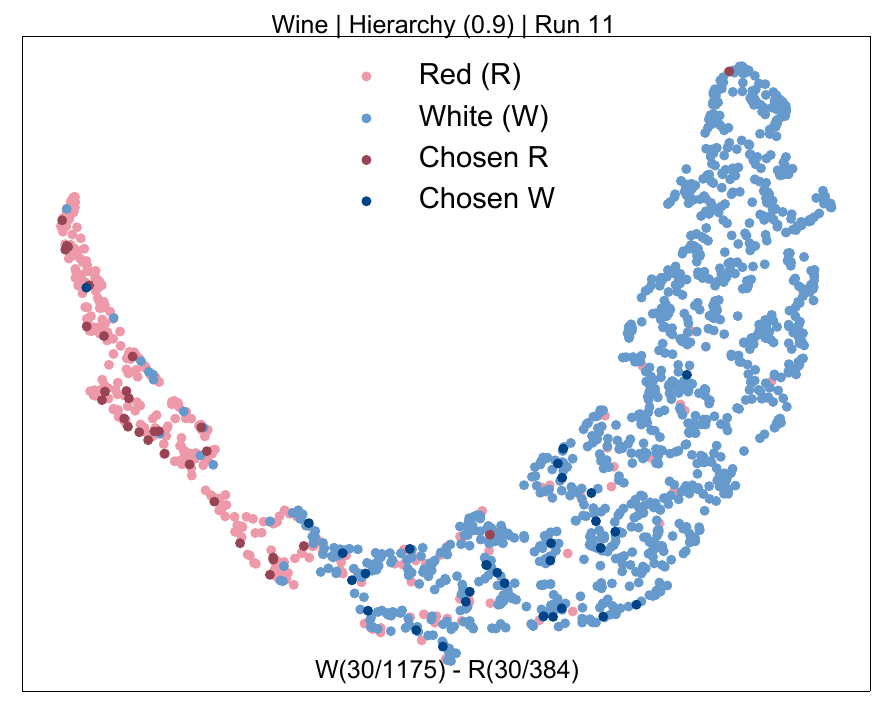}
         \caption{ }
         \label{supfig:umap_hierarchy_wine}
     \end{subfigure}
     \hfill
     \begin{subfigure}[b]{0.32\textwidth}
         \centering
         \includegraphics[width=\textwidth]{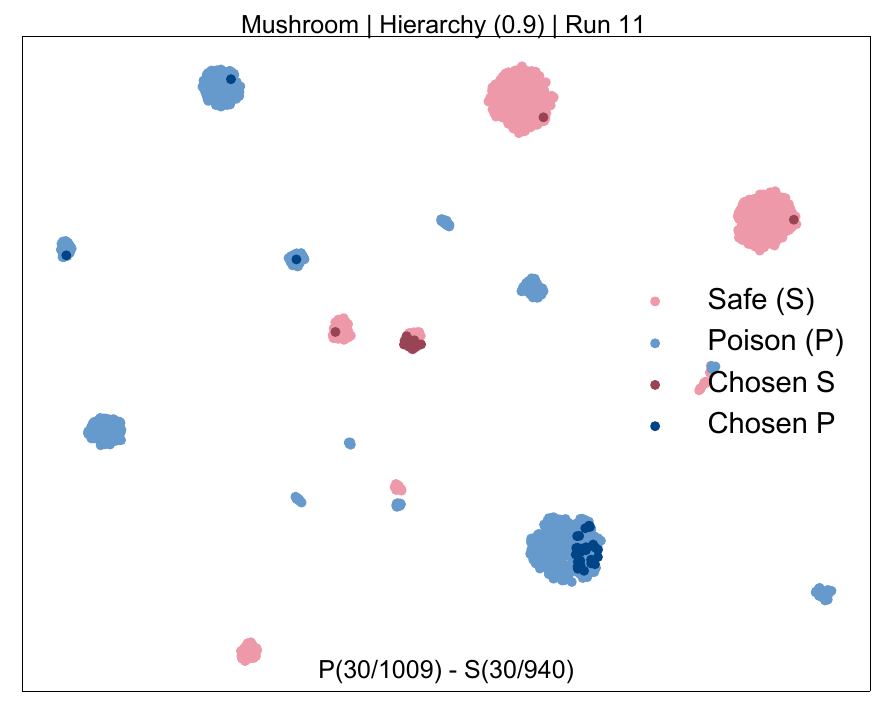}
         \caption{ }
         \label{supfig:umap_hierarchy_mushroom}
     \end{subfigure}
     \hfill
     \begin{subfigure}[b]{0.32\textwidth}
         \centering
         \includegraphics[width=\textwidth]{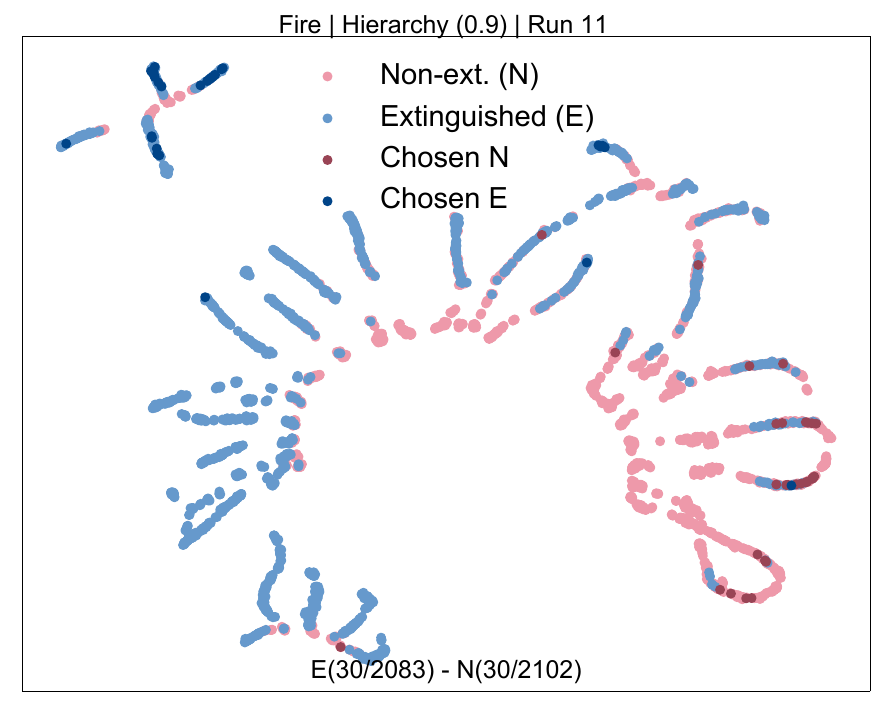}
         \caption{ }
         \label{supfig:umap_hierarchy_fire_random}
     \end{subfigure}
     \hfill
     \begin{subfigure}[b]{0.32\textwidth}
         \centering
         \includegraphics[width=\textwidth]{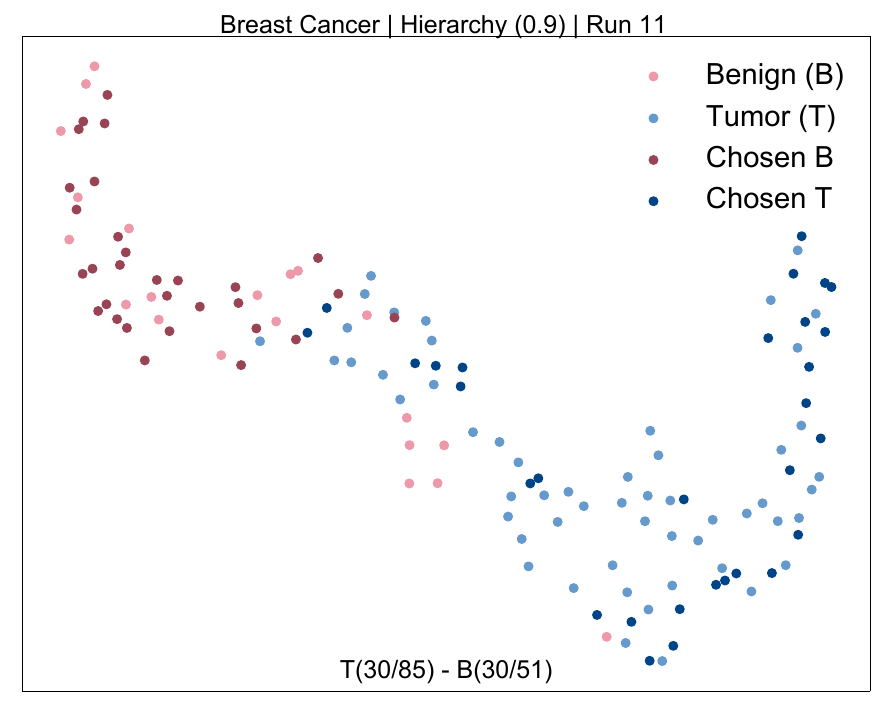}
         \caption{ }
         \label{supfig:umap_hierarchy_breast}
     \end{subfigure}
     \hfill
     \begin{subfigure}[b]{0.32\textwidth}
         \centering
         \includegraphics[width=\textwidth]{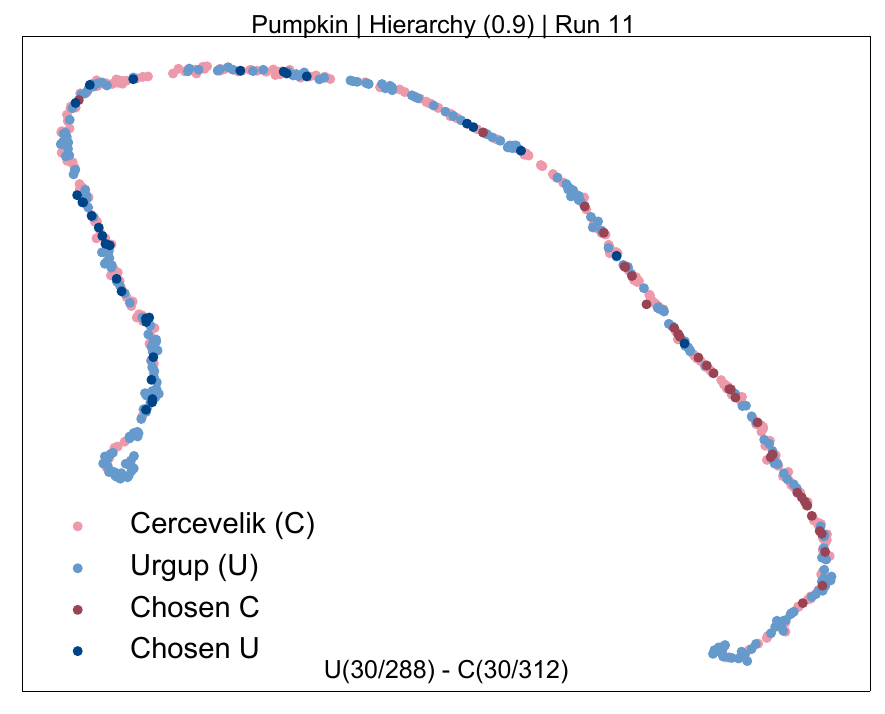}
         \caption{ }
         \label{supfig:umap_hierarchy_pumpkin}
     \end{subfigure}
     \hfill
     \begin{subfigure}[b]{0.32\textwidth}
         \centering
         \includegraphics[width=\textwidth]{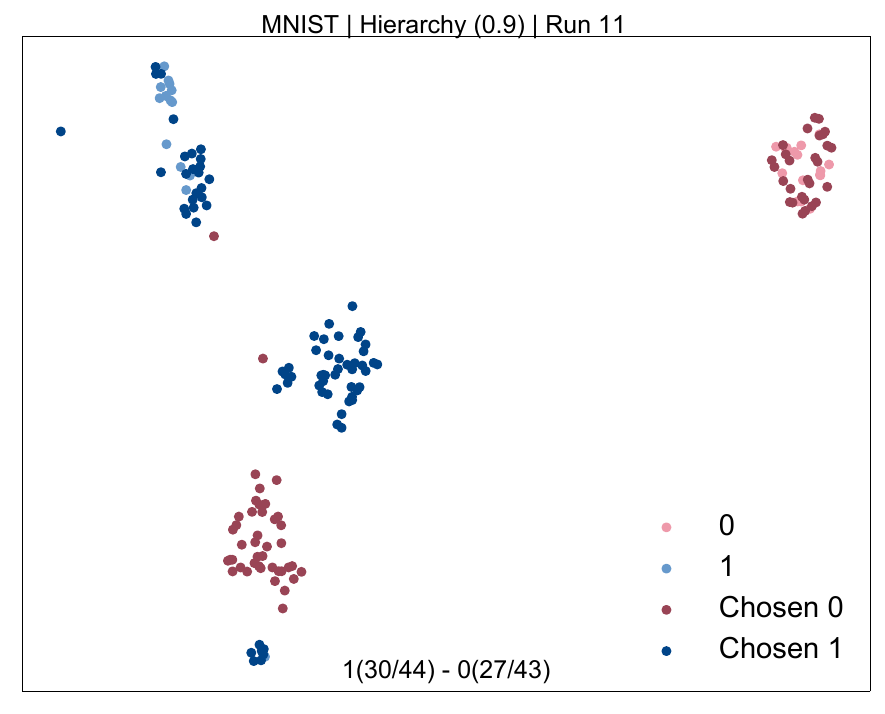}
         \caption{ }
         \label{supfig:umap_hierarchy_mnist}
     \end{subfigure}
     \hfill
     \begin{subfigure}[b]{0.32\textwidth}
         \centering
         \includegraphics[width=\textwidth]{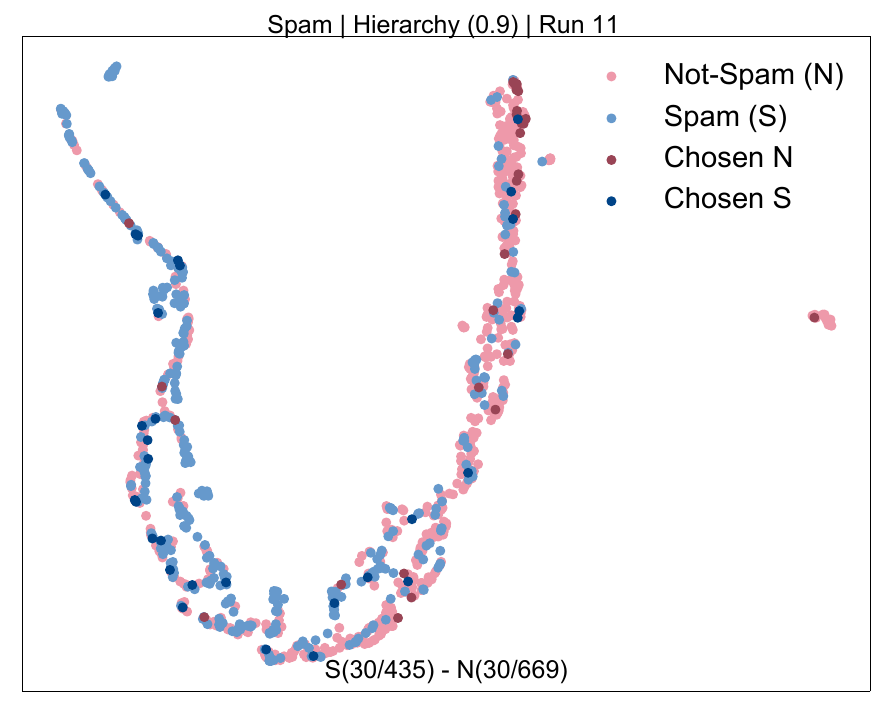}
         \caption{ }
         \label{supfig:umap_hierarchy_spam}
     \end{subfigure}
     \hfill
     \begin{subfigure}[b]{0.32\textwidth}
         \centering
         \includegraphics[width=\textwidth]{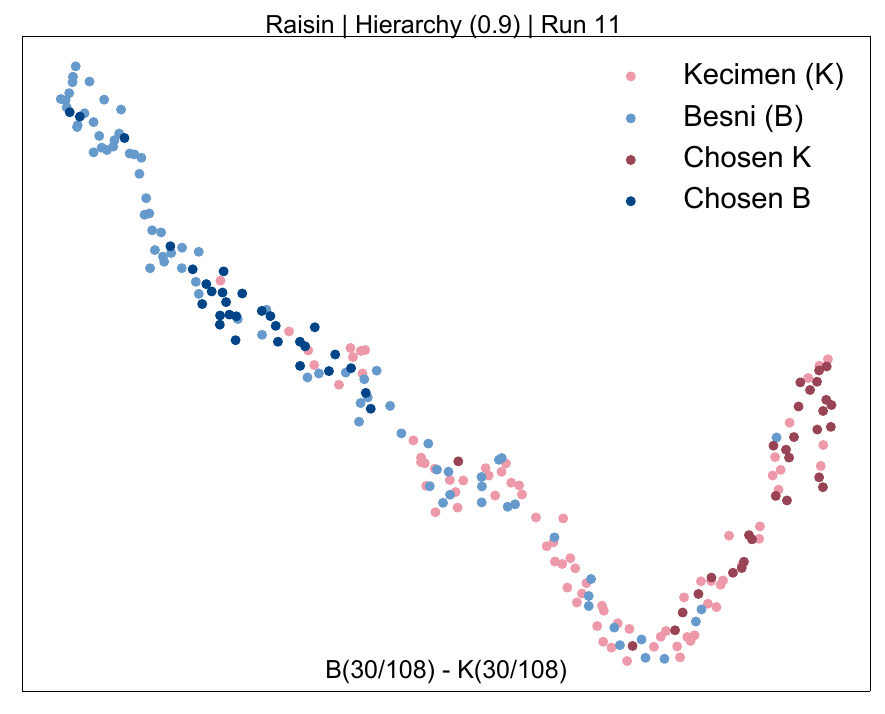}
         \caption{ }
         \label{supfig:umap_hierarchy_raisin}
     \end{subfigure}
     \hfill
     \begin{subfigure}[b]{0.32\textwidth}
         \centering
         \includegraphics[width=\textwidth]{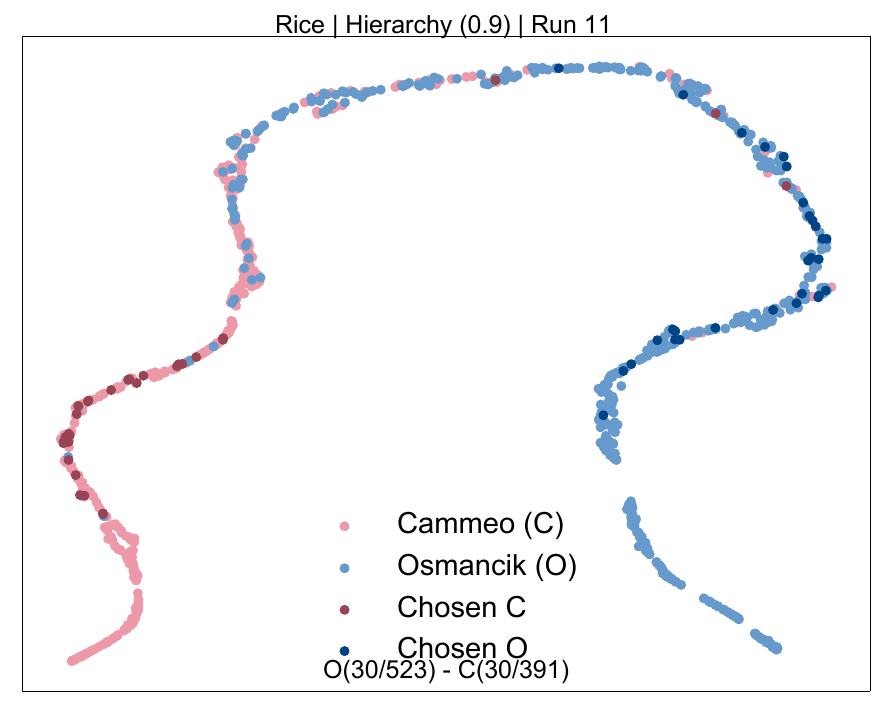}
         \caption{ }
         \label{supfig:umap_hierarchy_rice}
     \end{subfigure}
     \hfill
     \begin{subfigure}[b]{0.32\textwidth}
         \centering
         \includegraphics[width=\textwidth]{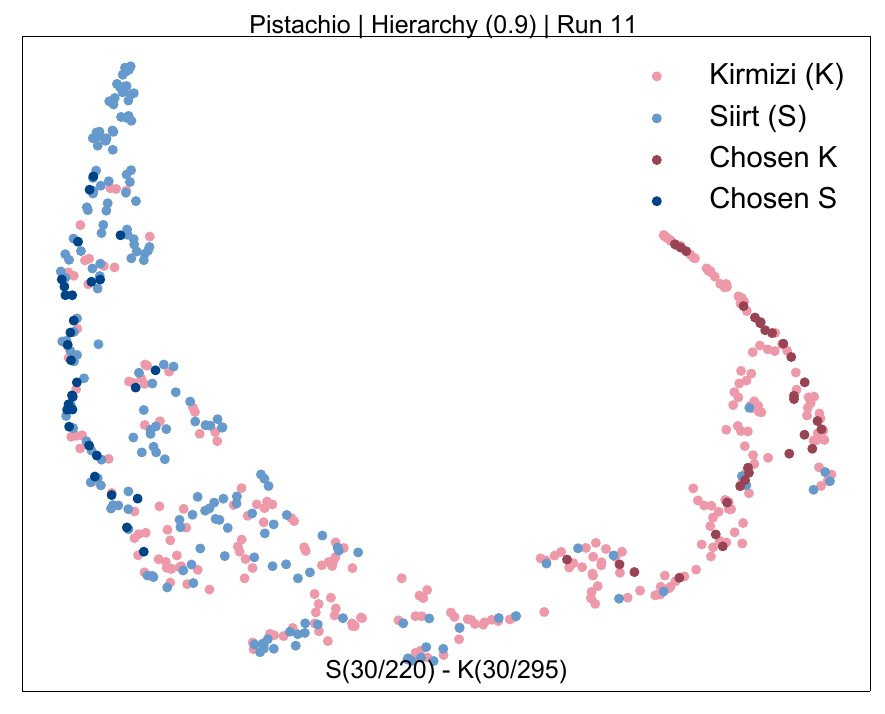}
         \caption{ }
         \label{supfig:umap_hierarchy_pistachio}
     \end{subfigure}
     \hfill
     \begin{subfigure}[b]{0.32\textwidth}
         \centering
         \includegraphics[width=\textwidth]{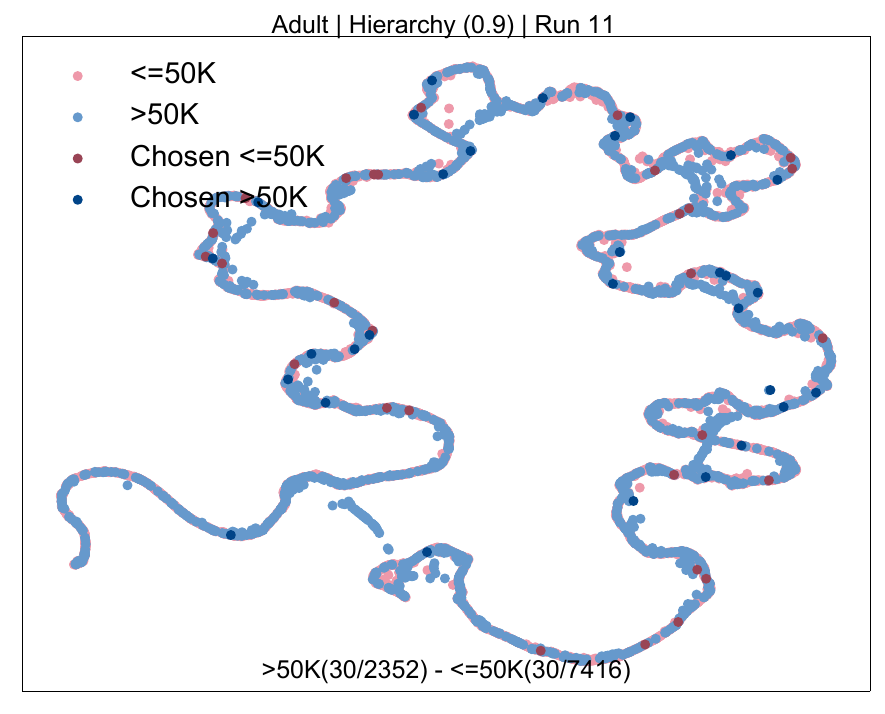}
         \caption{ }
         \label{supfig:umap_hierarchy_adult}
     \end{subfigure}
\caption{\textbf{Impact of hierarchy bias induction on the UMAP latent space.
} 
Samples selected by hierarchy bias ($b=0.9$) highlighted on the respective latent UMAP space of the labeled train set for each of the 11 datasets: (a) wine, (b) mushroom, (c) fire, (d) breast cancer, (e) pumpkin, (f) MNIST, (g) spam, (h) raisin, (i) rice, (j) pistachio, and (k) adult. Results are shown for run 11 (arbitrarily chosen).} 
\label{supfig:umap_hierarchy_all}
\end{figure}

\begin{figure}[th!]
     \begin{subfigure}[b]{0.32\textwidth}
         \centering
         \includegraphics[width=\textwidth]{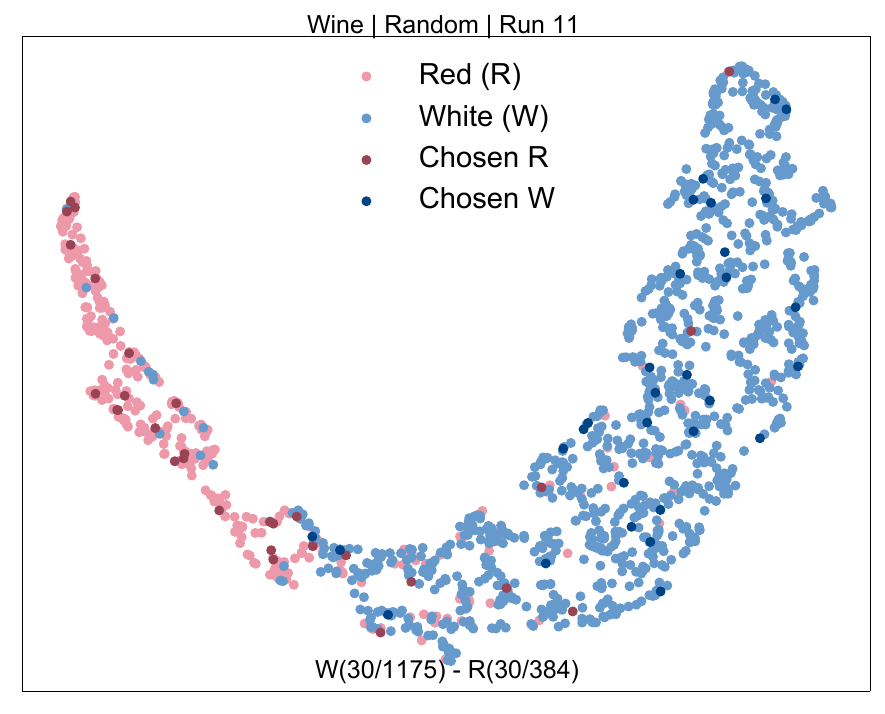}
         \caption{ }
         \label{supfig:umap_random_wine}
     \end{subfigure}
     \hfill
     \begin{subfigure}[b]{0.32\textwidth}
         \centering
         \includegraphics[width=\textwidth]{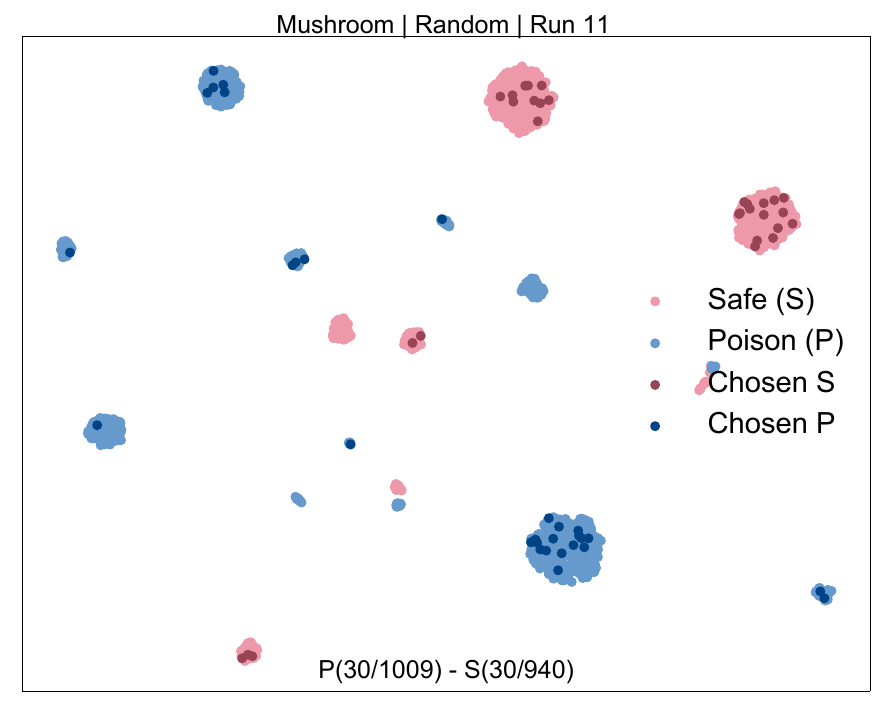}
         \caption{ }
         \label{supfig:umap_random_mushroom}
     \end{subfigure}
     \hfill
     \begin{subfigure}[b]{0.32\textwidth}
         \centering
         \includegraphics[width=\textwidth]{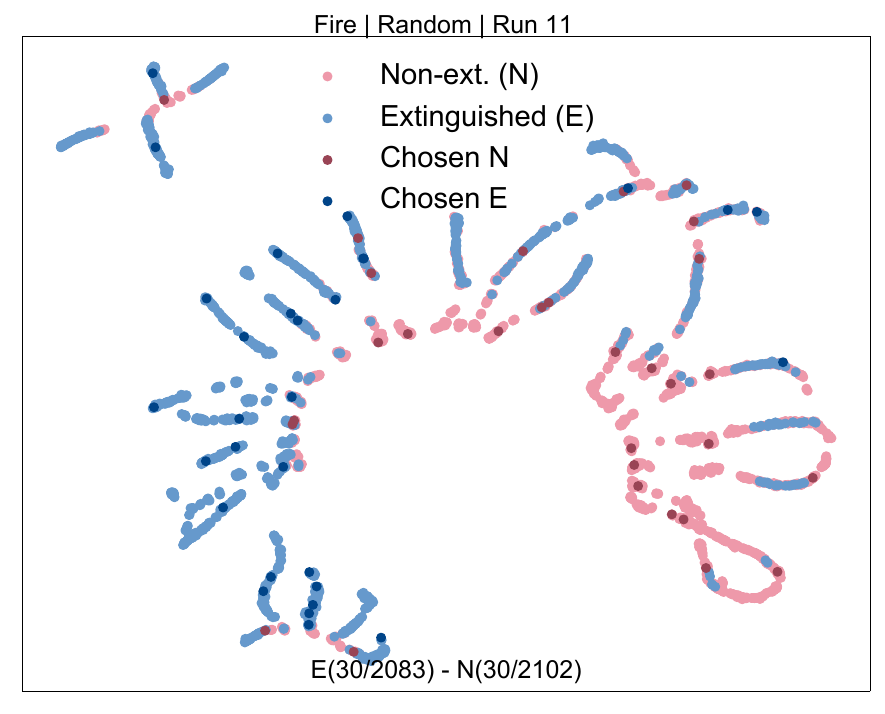}
         \caption{ }
         \label{supfig:umap_random_fire}
     \end{subfigure}
     \hfill
     \begin{subfigure}[b]{0.32\textwidth}
         \centering
         \includegraphics[width=\textwidth]{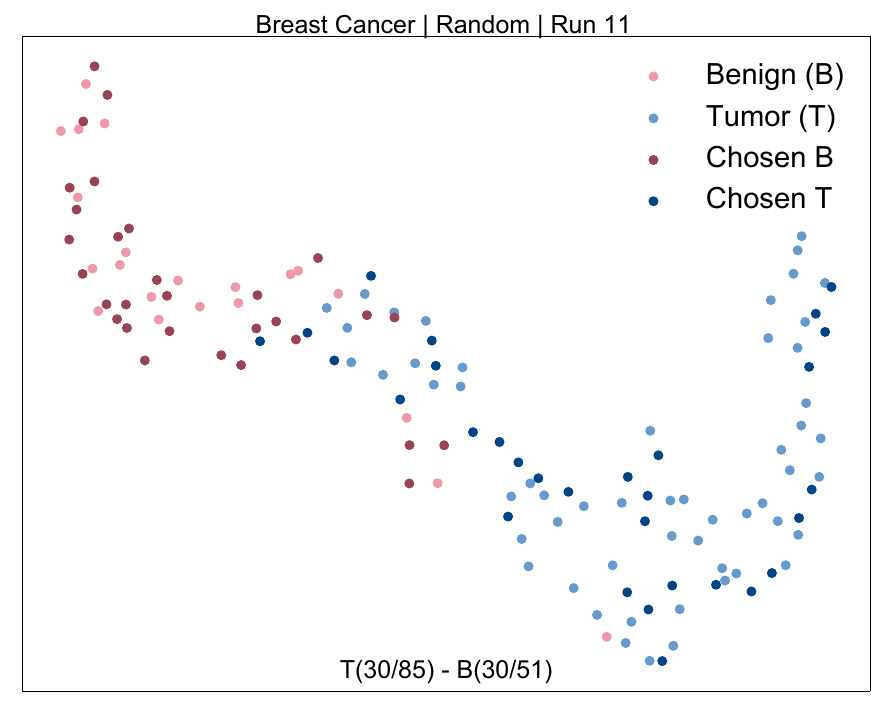}
         \caption{ }
         \label{supfig:umap_random_breast}
     \end{subfigure}
     \hfill
     \begin{subfigure}[b]{0.32\textwidth}
         \centering
         \includegraphics[width=\textwidth]{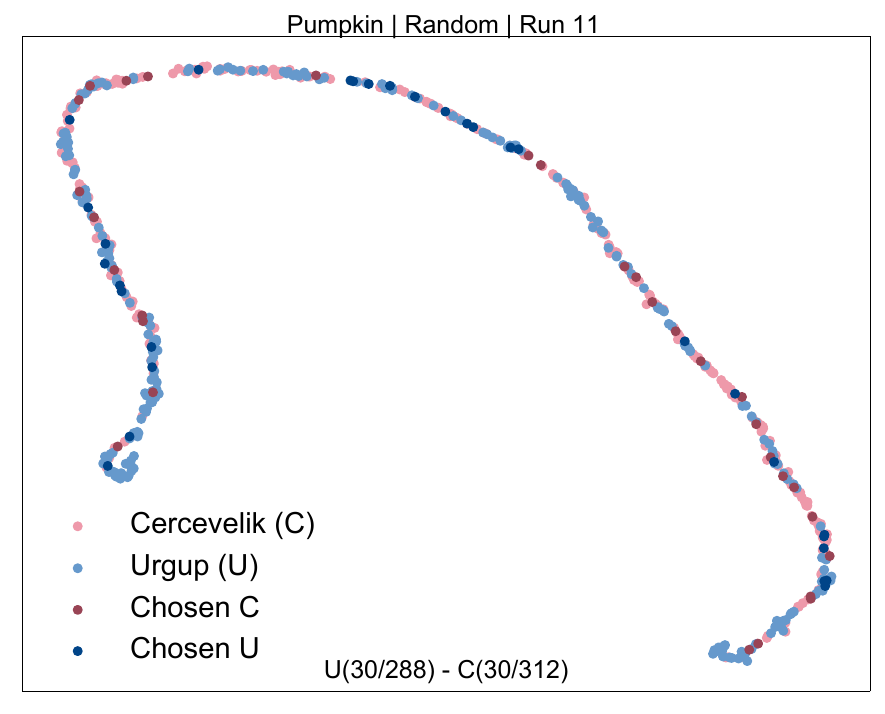}
         \caption{ }
         \label{supfig:umap_random_pumpkin}
     \end{subfigure}
     \hfill
     \begin{subfigure}[b]{0.32\textwidth}
         \centering
         \includegraphics[width=\textwidth]{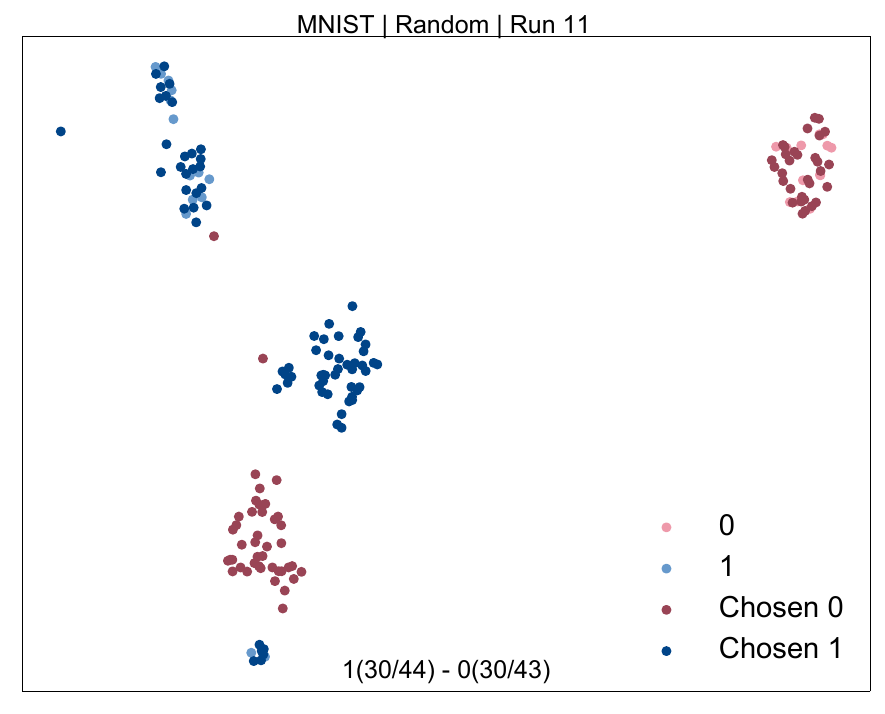}
         \caption{ }
         \label{supfig:umap_random_mnist}
     \end{subfigure}
     \hfill
     \begin{subfigure}[b]{0.32\textwidth}
         \centering
         \includegraphics[width=\textwidth]{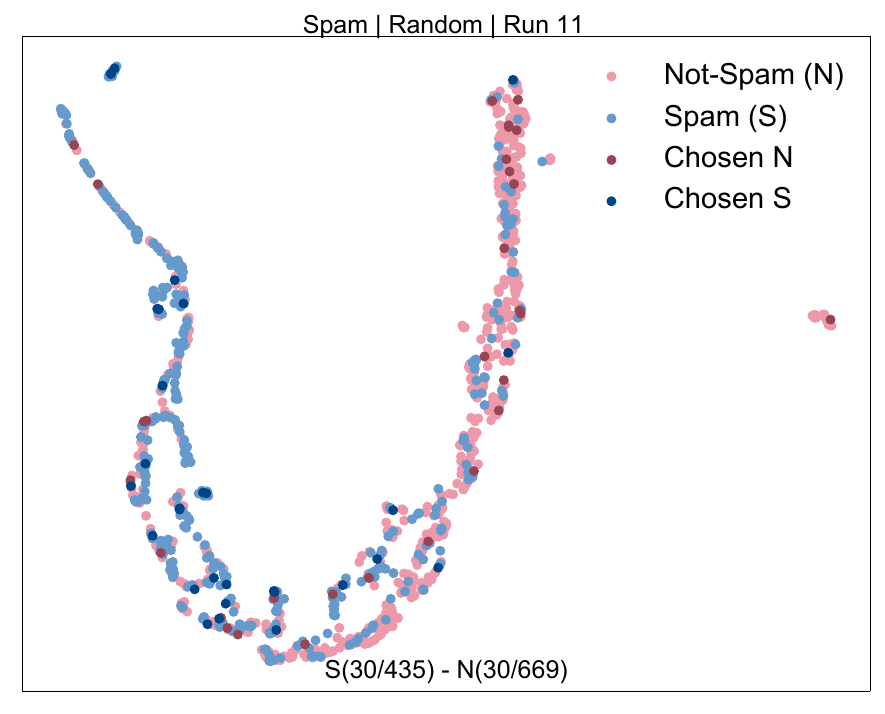}
         \caption{ }
         \label{supfig:umap_random_spam}
     \end{subfigure}
     \hfill
     \begin{subfigure}[b]{0.32\textwidth}
         \centering
         \includegraphics[width=\textwidth]{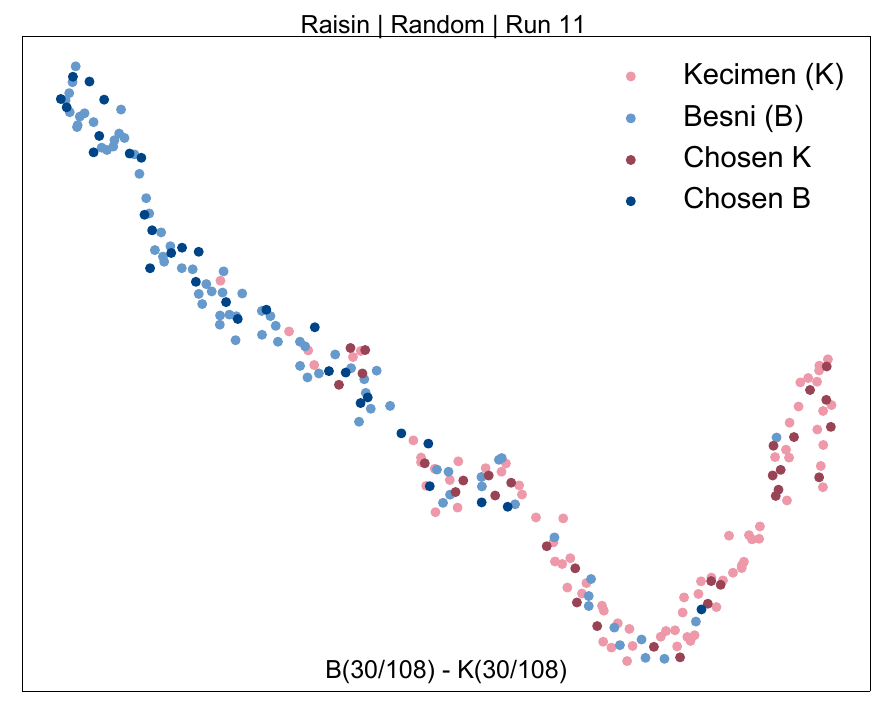}
         \caption{ }
         \label{supfig:umap_random_raisin}
     \end{subfigure}
     \hfill
     \begin{subfigure}[b]{0.32\textwidth}
         \centering
         \includegraphics[width=\textwidth]{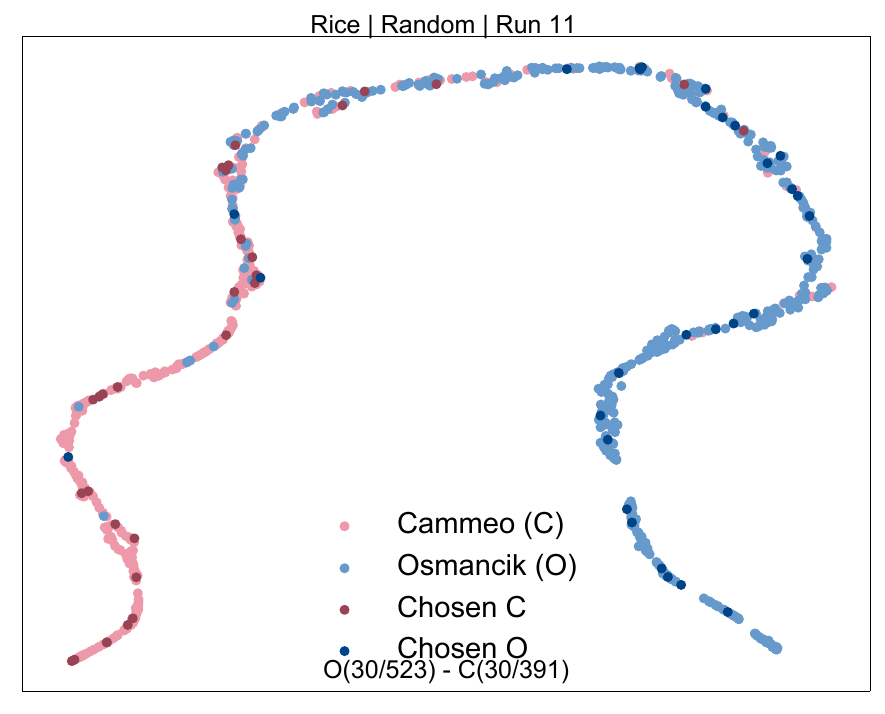}
         \caption{ }
         \label{supfig:umap_random_rice}
     \end{subfigure}
     \hfill
     \begin{subfigure}[b]{0.32\textwidth}
         \centering
         \includegraphics[width=\textwidth]{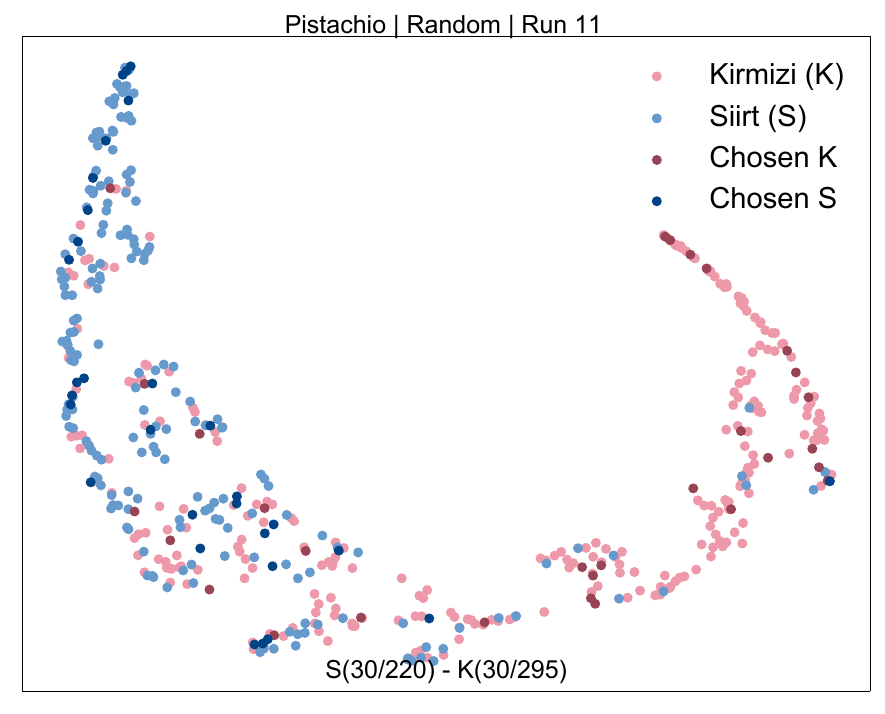}
         \caption{ }
         \label{supfig:umap_random_pistachio}
     \end{subfigure}
     \hfill
     \begin{subfigure}[b]{0.32\textwidth}
         \centering
         \includegraphics[width=\textwidth]{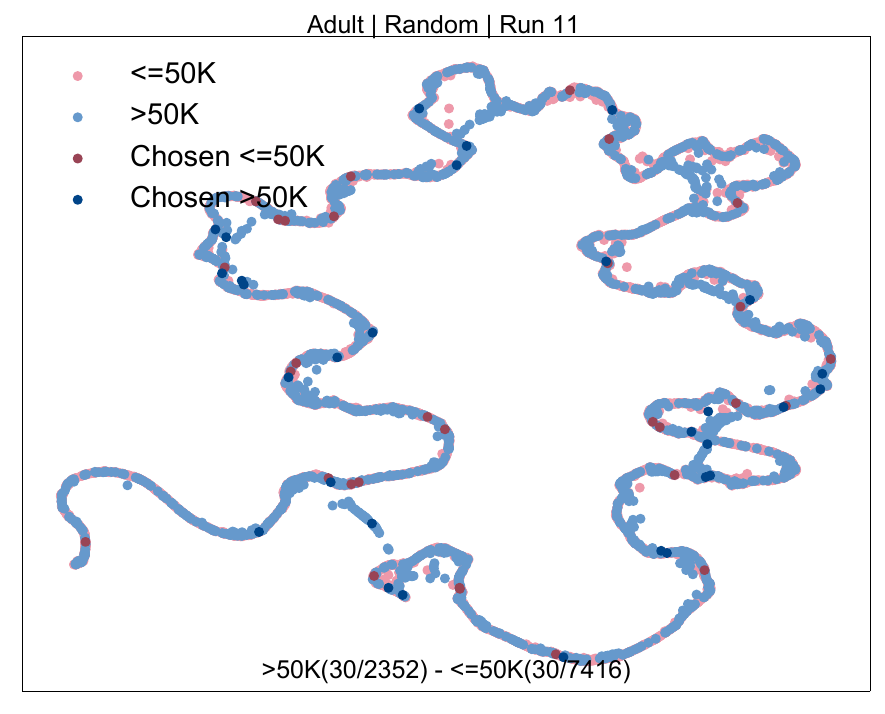}
         \caption{ }
         \label{supfig:umap_random_adult}
     \end{subfigure}
\caption{\textbf{Impact of random subsampling on the UMAP latent space. } 
Samples selected by random subsampling highlighted on the respective latent UMAP space of the labeled train set for each of the 11 datasets: (a) wine, (b) mushroom, (c) fire, (d) breast cancer, (e) pumpkin, (f) MNIST, (g) spam, (h) raisin, (i) rice, (j) pistachio, and (k) adult. Results are shown for run 11 (arbitrarily chosen).} 
\label{supfig:umap_random_all}
\end{figure}

\begin{figure}[th!]
\centering
     \begin{subfigure}[b]{0.32\textwidth}
         \centering
         \includegraphics[width=\textwidth]{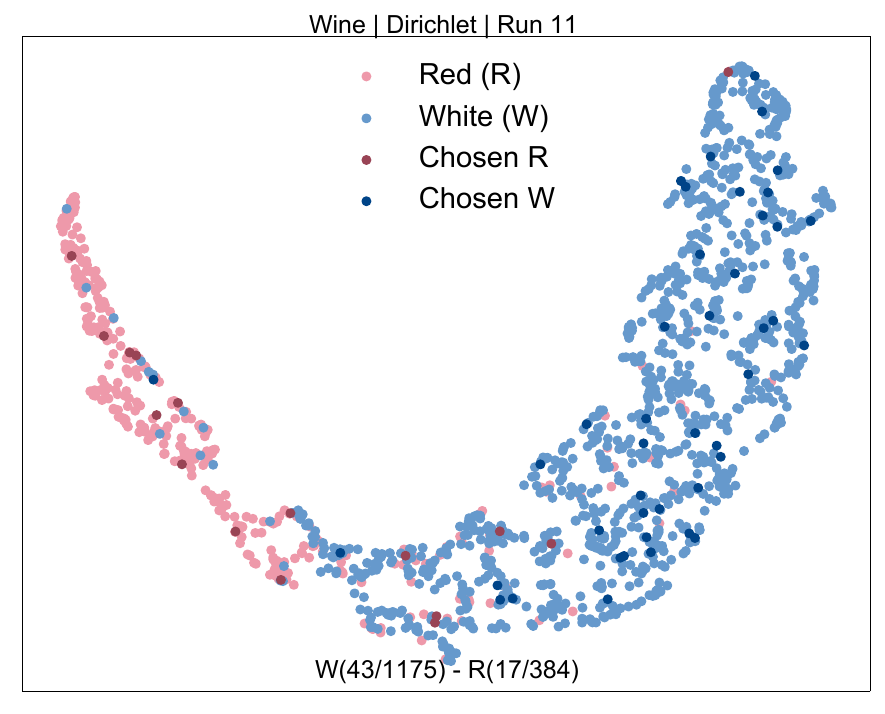}
         \caption{ }
         \label{supfig:umap_dirichlet_wine}
     \end{subfigure}
     \hfill
     \begin{subfigure}[b]{0.32\textwidth}
         \centering
         \includegraphics[width=\textwidth]{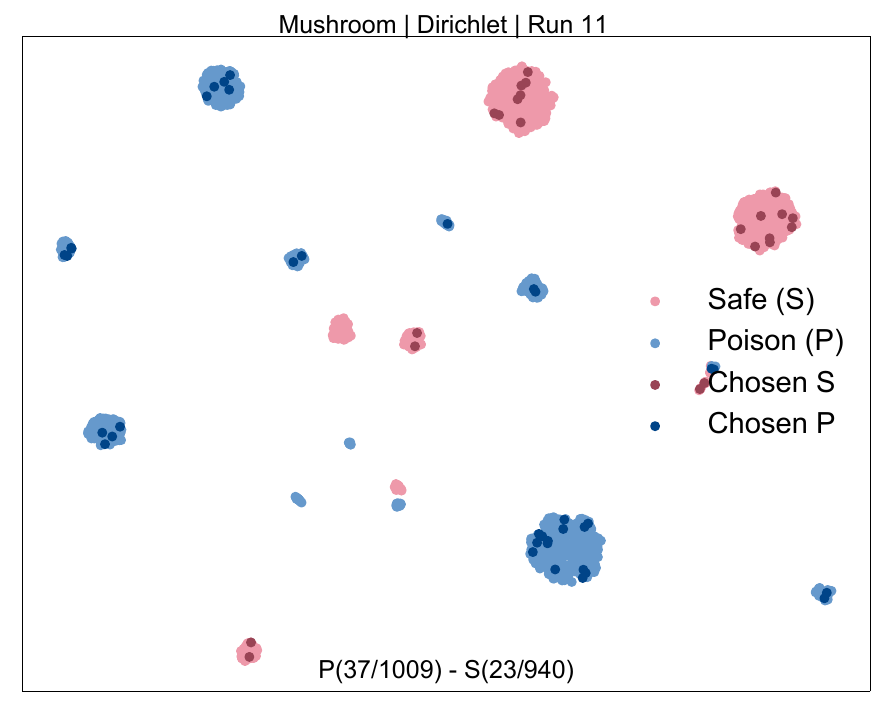}
         \caption{ }
         \label{supfig:umap_dirichlet_mushroom}
     \end{subfigure}
     \hfill
     \begin{subfigure}[b]{0.32\textwidth}
         \centering
         \includegraphics[width=\textwidth]{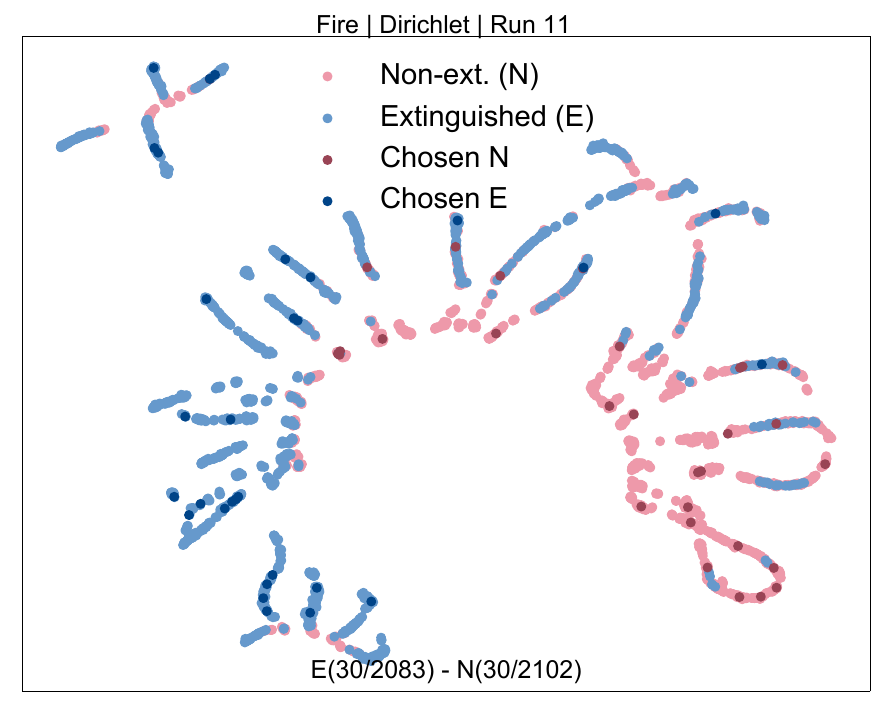}
         \caption{ }
         \label{supfig:umap_dirichlet_fire}
     \end{subfigure}
     \hfill
     \begin{subfigure}[b]{0.32\textwidth}
         \centering
         \includegraphics[width=\textwidth]{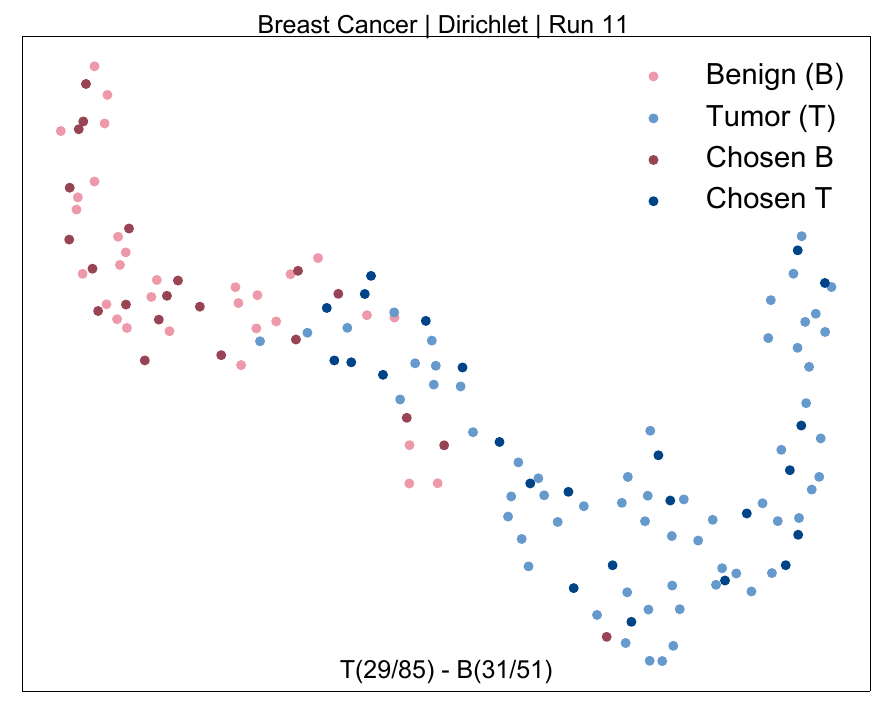}
         \caption{ }
         \label{supfig:umap_dirichlet_breast}
     \end{subfigure}
     \hfill
     \begin{subfigure}[b]{0.32\textwidth}
         \centering
         \includegraphics[width=\textwidth]{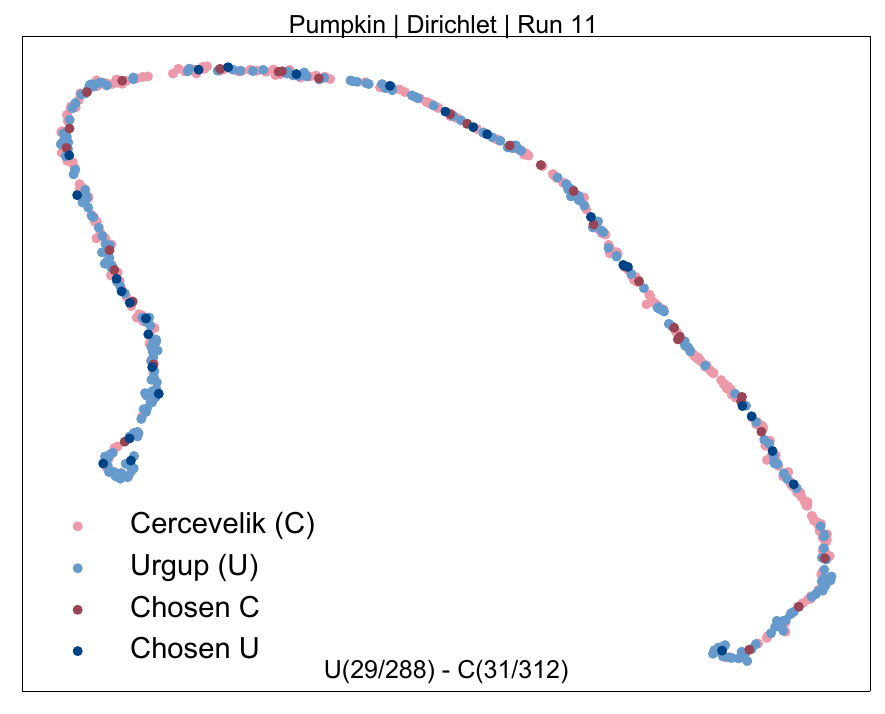}
         \caption{ }
         \label{supfig:umap_dirichlet_pumpkin}
     \end{subfigure}
     \hfill
     \begin{subfigure}[b]{0.32\textwidth}
         \centering
         \includegraphics[width=\textwidth]{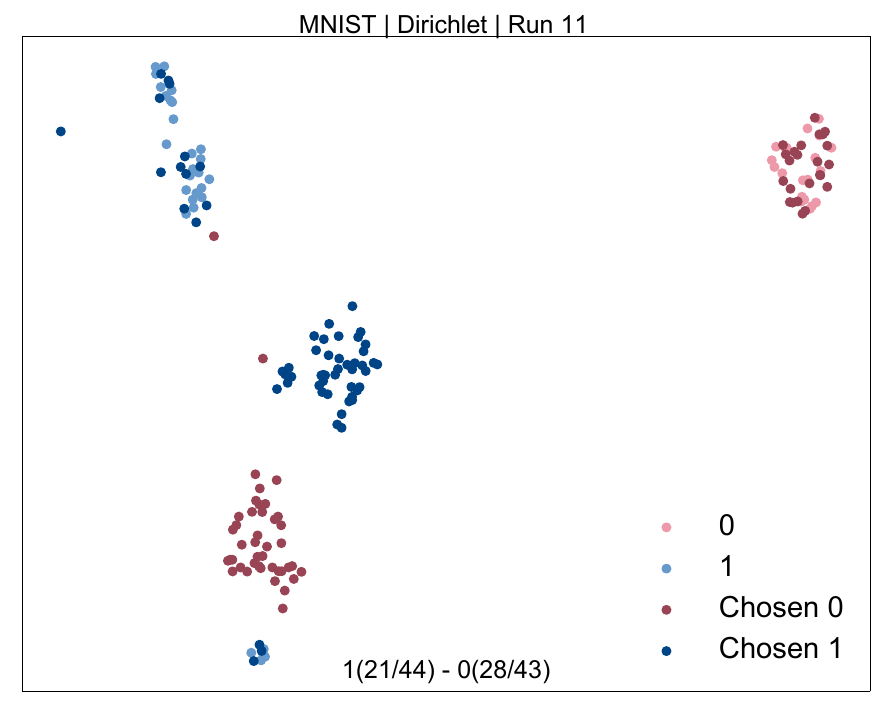}
         \caption{ }
         \label{supfig:umap_dirichlet_mnist}
     \end{subfigure}
     \hfill
     \begin{subfigure}[b]{0.32\textwidth}
         \centering
         \includegraphics[width=\textwidth]{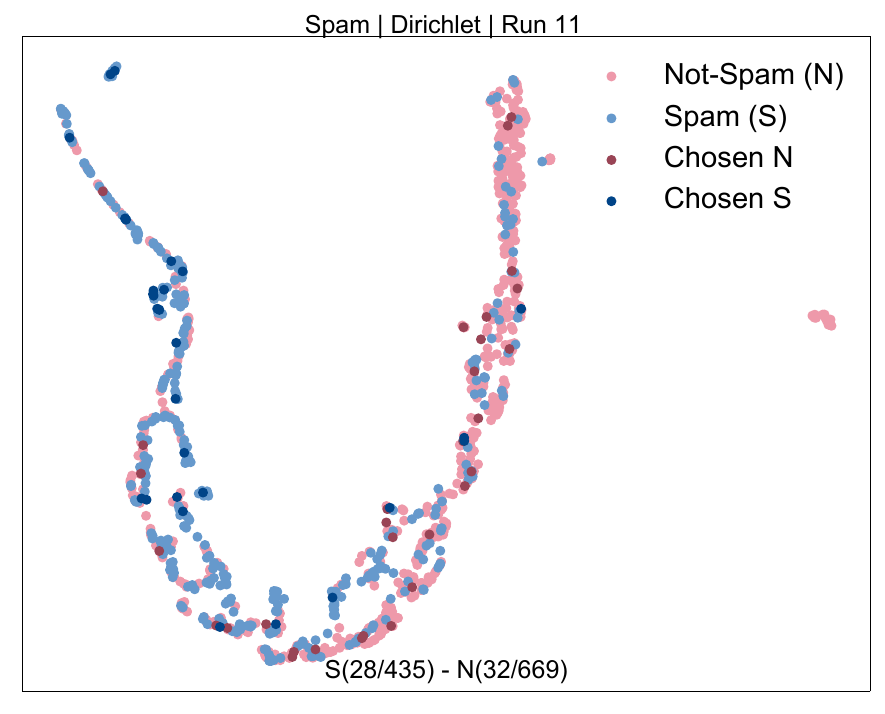}
         \caption{ }
         \label{supfig:umap_dirichlet_spam}
     \end{subfigure}
     \hfill
     \begin{subfigure}[b]{0.32\textwidth}
         \centering
         \includegraphics[width=\textwidth]{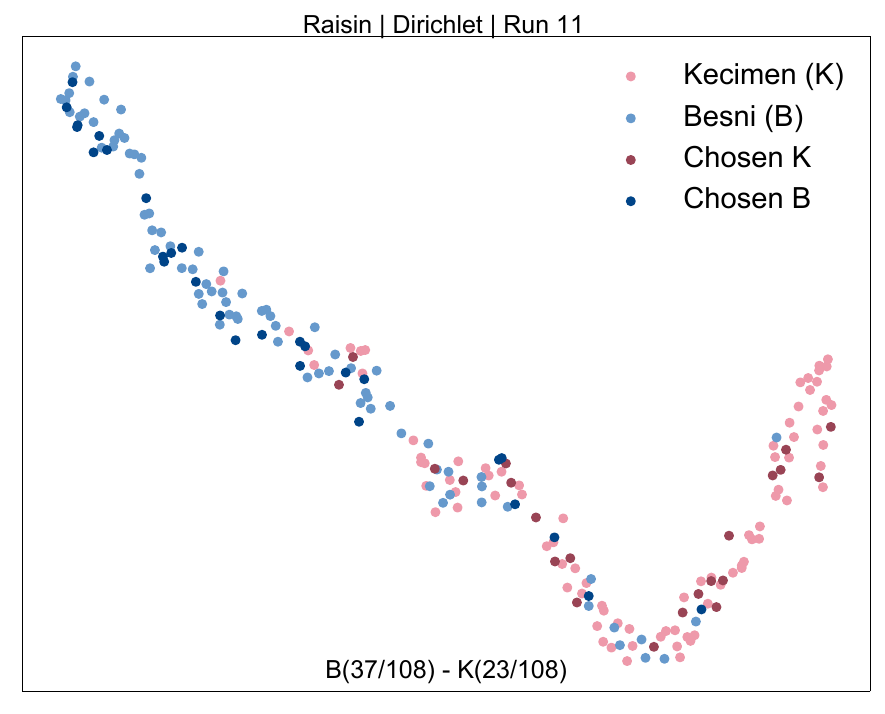}
         \caption{ }
         \label{supfig:umap_dirichlet_raisin}
     \end{subfigure}
     \hfill
     \begin{subfigure}[b]{0.32\textwidth}
         \centering
         \includegraphics[width=\textwidth]{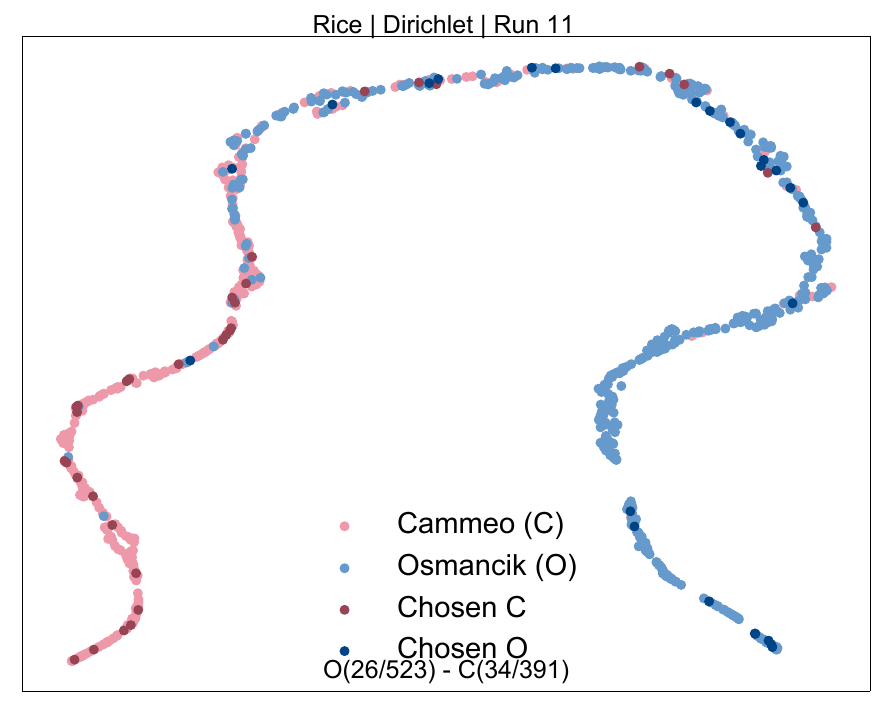}
         \caption{ }
         \label{supfig:umap_dirichlet_rice}
     \end{subfigure}
     \hfill
     \begin{subfigure}[b]{0.32\textwidth}
         \centering
         \includegraphics[width=\textwidth]{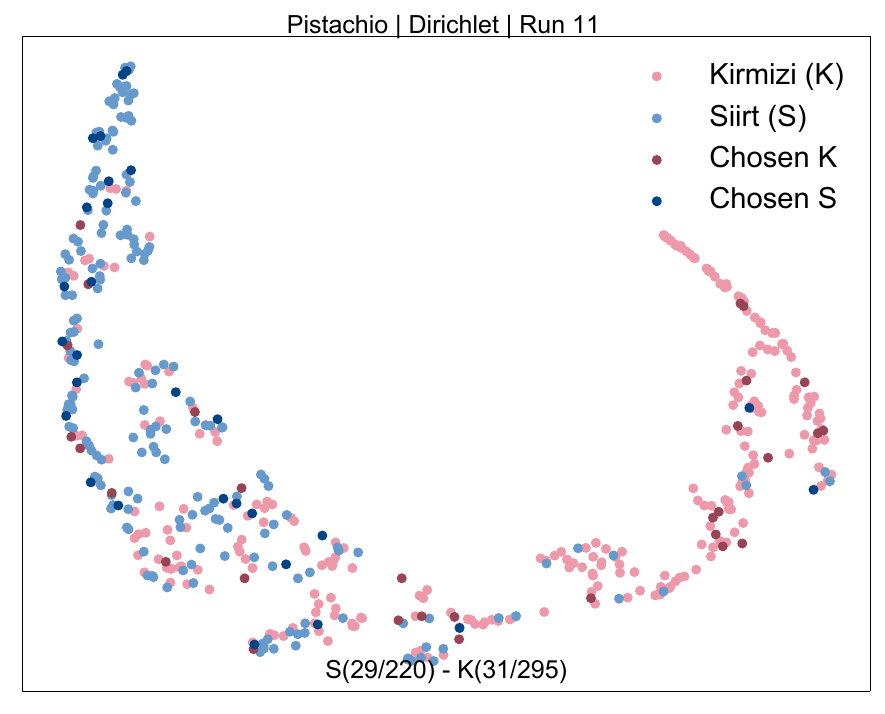}
         \caption{ }
         \label{supfig:umap_dirichlet_pistachio}
     \end{subfigure}
     \hfill
     \begin{subfigure}[b]{0.32\textwidth}
         \centering
         \includegraphics[width=\textwidth]{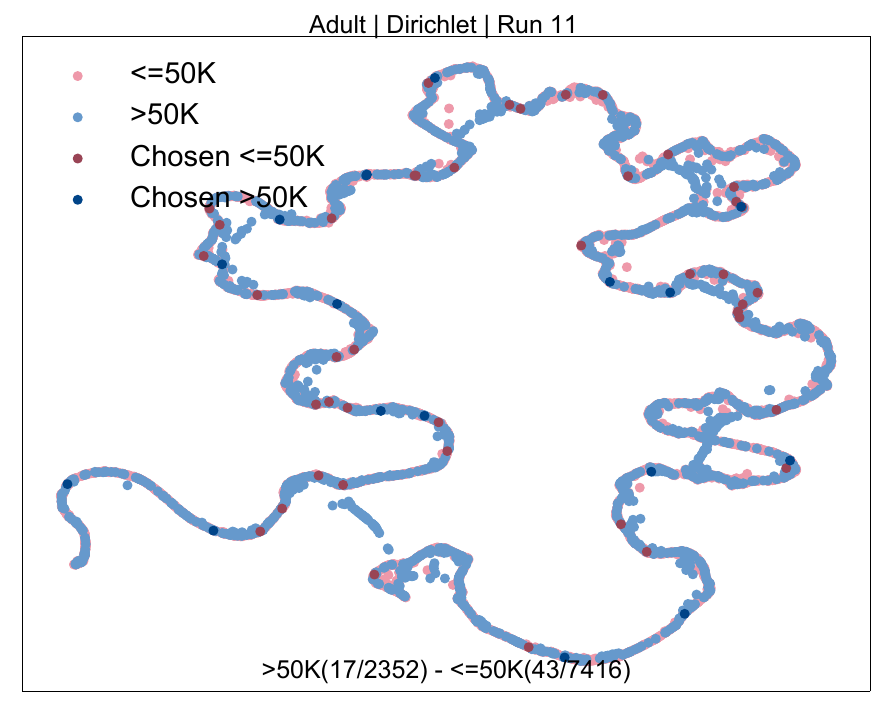}
         \caption{ }
         \label{supfig:umap_dirichlet_adult}
     \end{subfigure}
\caption{\textbf{Impact of Dirichlet bias induction on the UMAP latent space. } 
Samples selected by Dirichlet bias highlighted on the respective latent UMAP space of the labeled train set for each of the 11 datasets: (a) wine, (b) mushroom, (c) fire, (d) breast cancer, (e) pumpkin, (f) MNIST, (g) spam, (h) raisin, (i) rice, (j) pistachio, and (k) adult. Results are shown for run 11 (arbitrarily chosen).} 
\label{supfig:umap_dirichlet_all}
\end{figure}

\begin{figure}[th!]
\centering
     \begin{subfigure}[b]{0.32\textwidth}
         \centering
         \includegraphics[width=\textwidth]{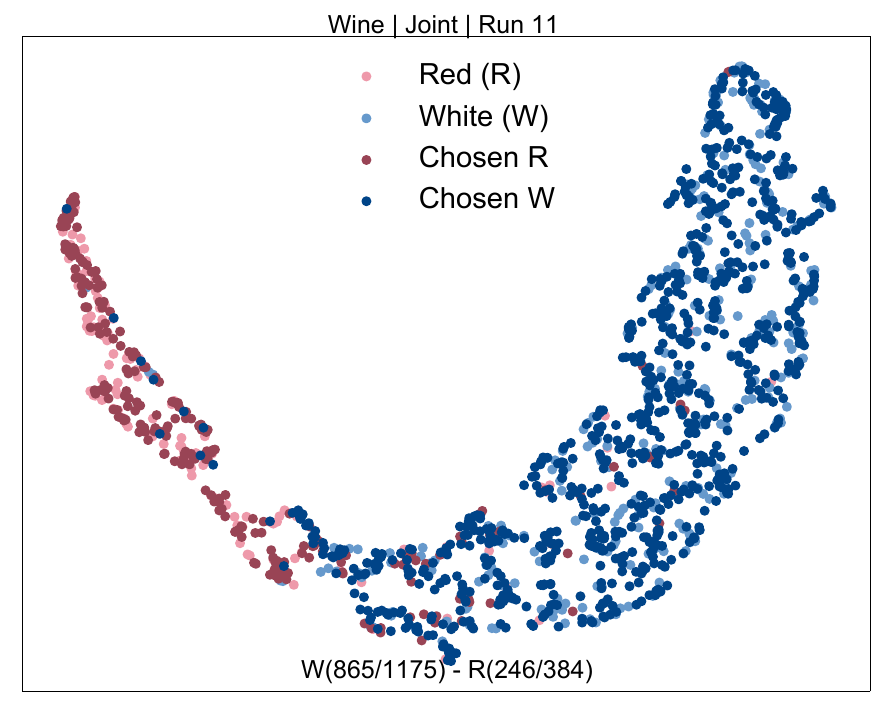}
         \caption{ }
         \label{supfig:umap_joint_wine}
     \end{subfigure}
     \hfill
     \begin{subfigure}[b]{0.32\textwidth}
         \centering
         \includegraphics[width=\textwidth]{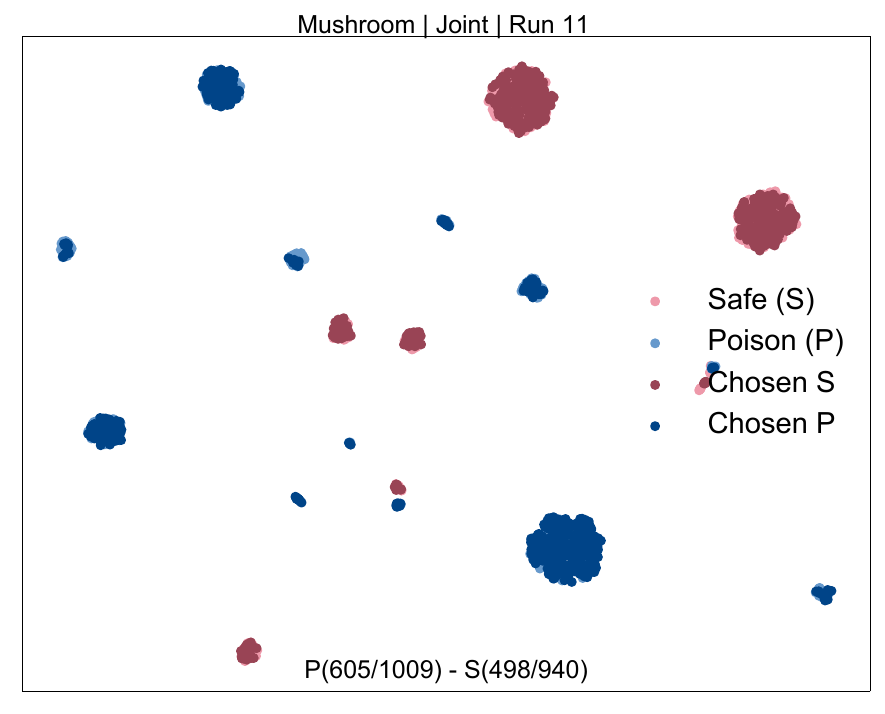}
         \caption{ }
         \label{supfig:umap_joint_mushroom}
     \end{subfigure}
     \hfill
     \begin{subfigure}[b]{0.32\textwidth}
         \centering
         \includegraphics[width=\textwidth]{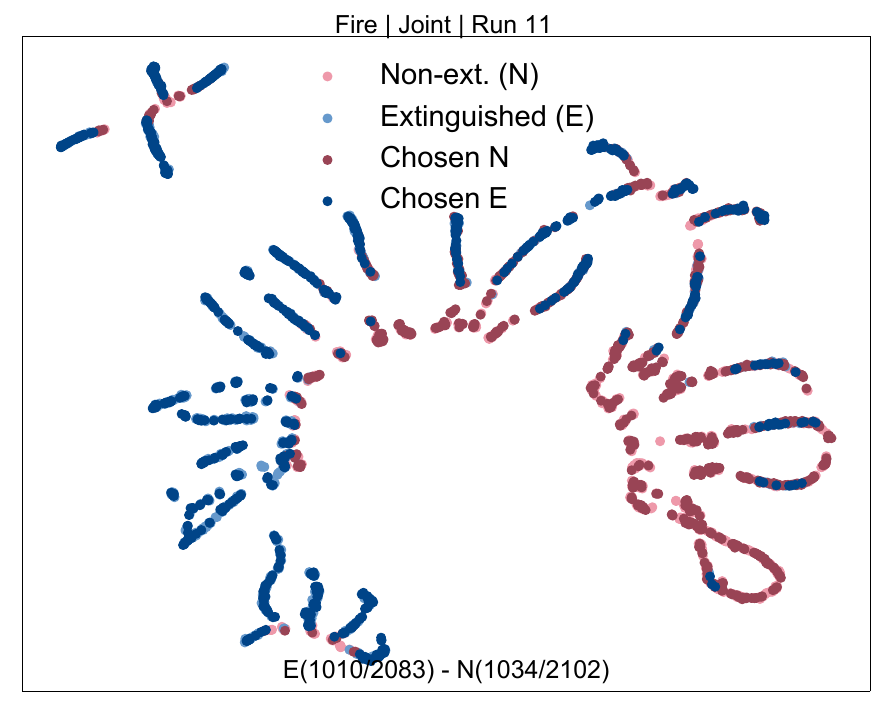}
         \caption{ }
         \label{supfig:umap_joint_fire}
     \end{subfigure}
     \hfill
     \begin{subfigure}[b]{0.32\textwidth}
         \centering
         \includegraphics[width=\textwidth]{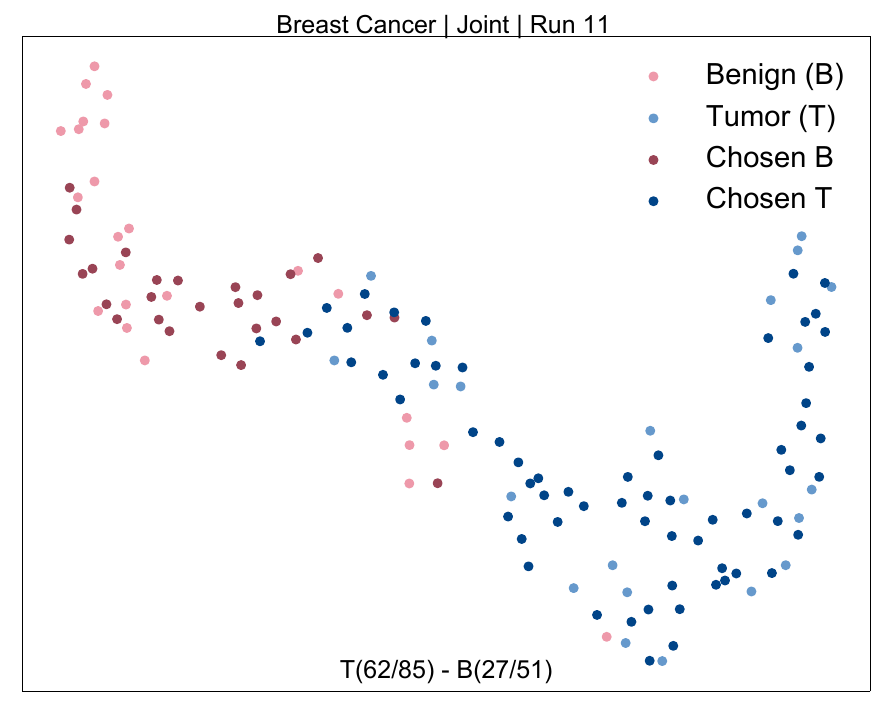}
         \caption{ }
         \label{supfig:umap_joint_breast}
     \end{subfigure}
     \hfill
     \begin{subfigure}[b]{0.32\textwidth}
         \centering
         \includegraphics[width=\textwidth]{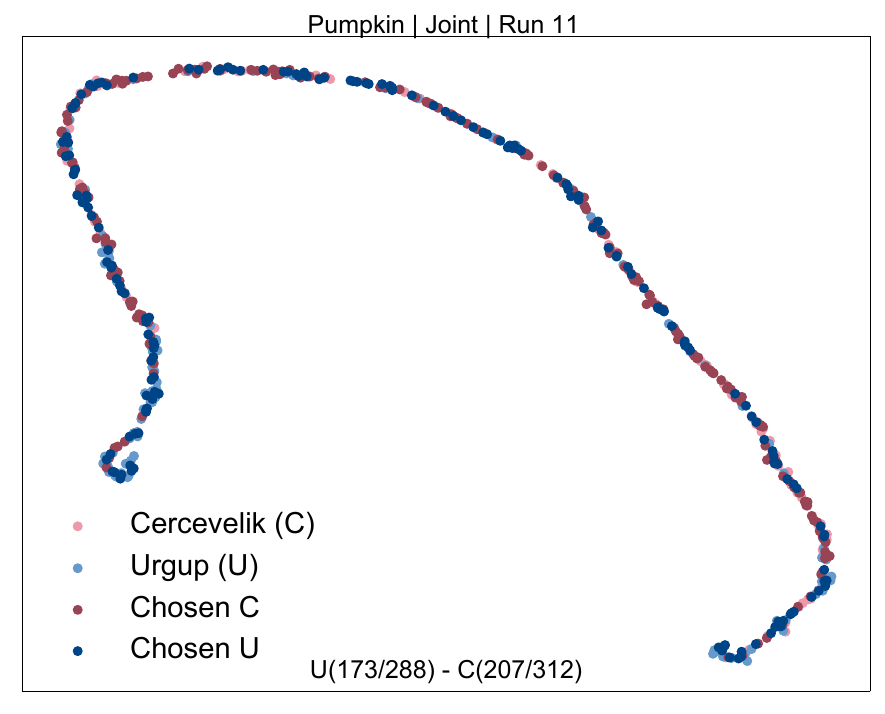}
         \caption{ }
         \label{supfig:umap_joint_pumpkin}
     \end{subfigure}
     \hfill
     \begin{subfigure}[b]{0.32\textwidth}
         \centering
         \includegraphics[width=\textwidth]{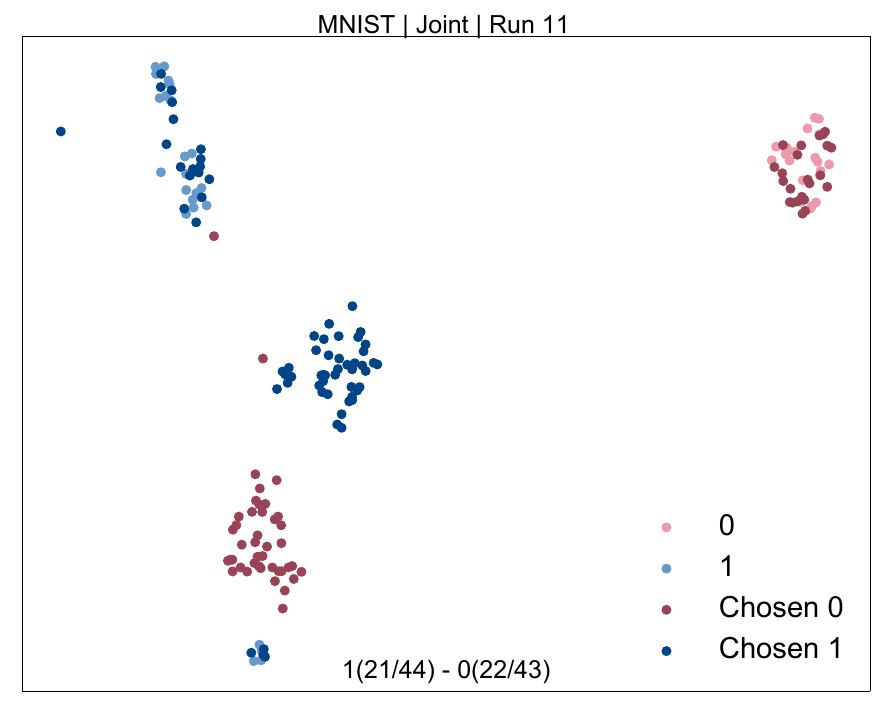}
         \caption{ }
         \label{supfig:umap_joint_mnist}
     \end{subfigure}
     \hfill
     \begin{subfigure}[b]{0.32\textwidth}
         \centering
         \includegraphics[width=\textwidth]{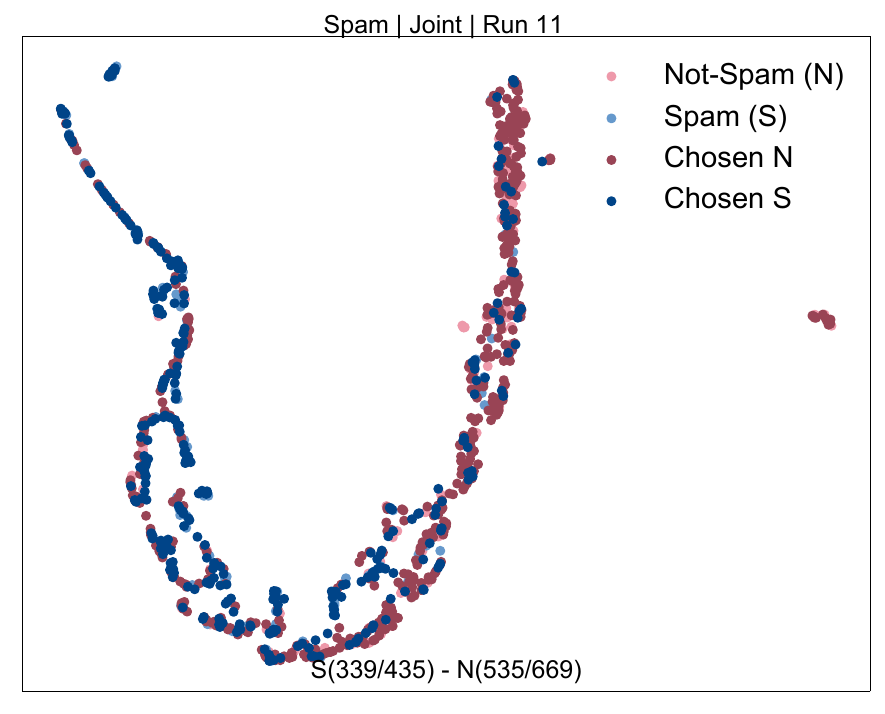}
         \caption{ }
         \label{supfig:umap_joint_spam}
     \end{subfigure}
     \hfill
     \begin{subfigure}[b]{0.32\textwidth}
         \centering
         \includegraphics[width=\textwidth]{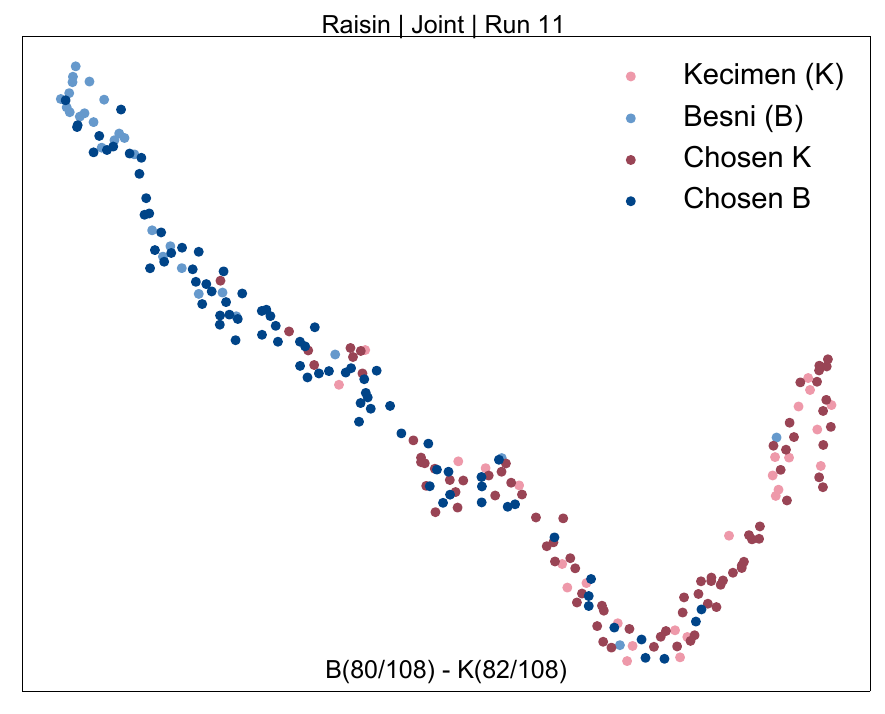}
         \caption{ }
         \label{supfig:umap_joint_raisin}
     \end{subfigure}
     \hfill
     \begin{subfigure}[b]{0.32\textwidth}
         \centering
         \includegraphics[width=\textwidth]{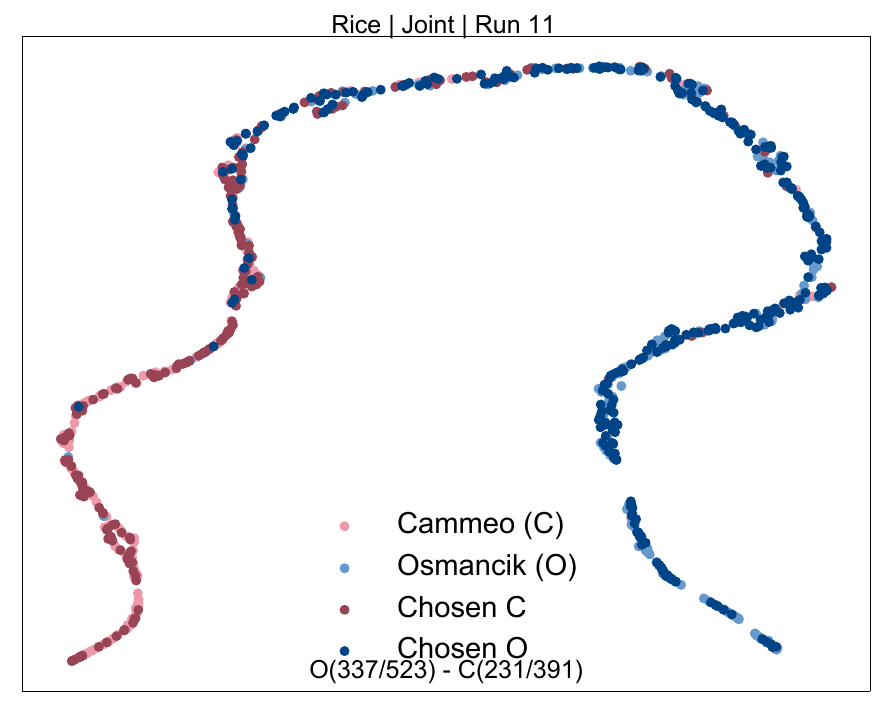}
         \caption{ }
         \label{supfig:umap_joint_rice}
     \end{subfigure}
     \hfill
     \begin{subfigure}[b]{0.32\textwidth}
         \centering
         \includegraphics[width=\textwidth]{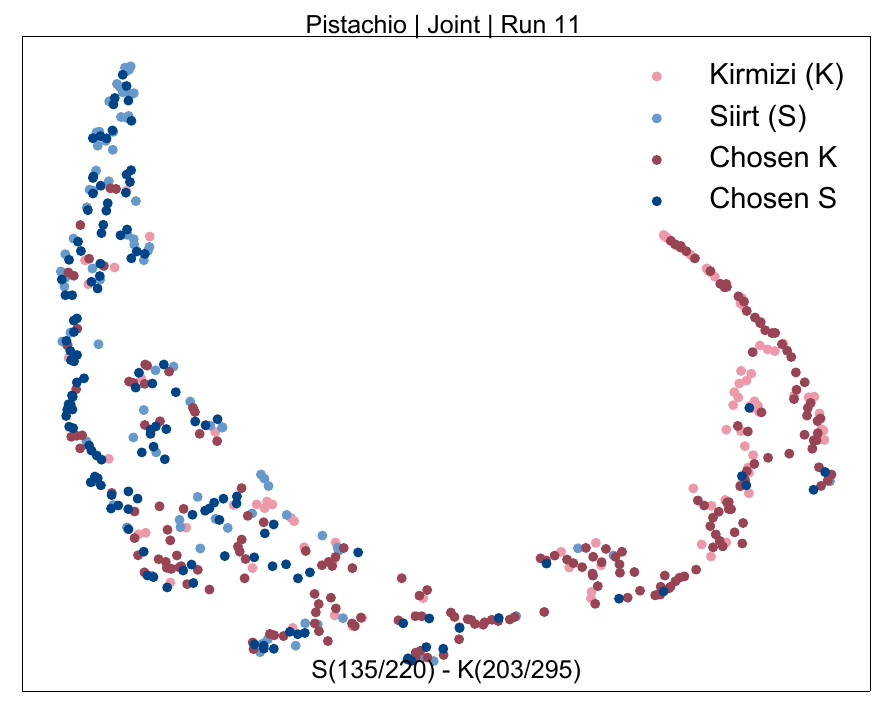}
         \caption{ }
         \label{supfig:umap_joint_pistachio}
     \end{subfigure}
     \hfill
     \begin{subfigure}[b]{0.32\textwidth}
         \centering
         \includegraphics[width=\textwidth]{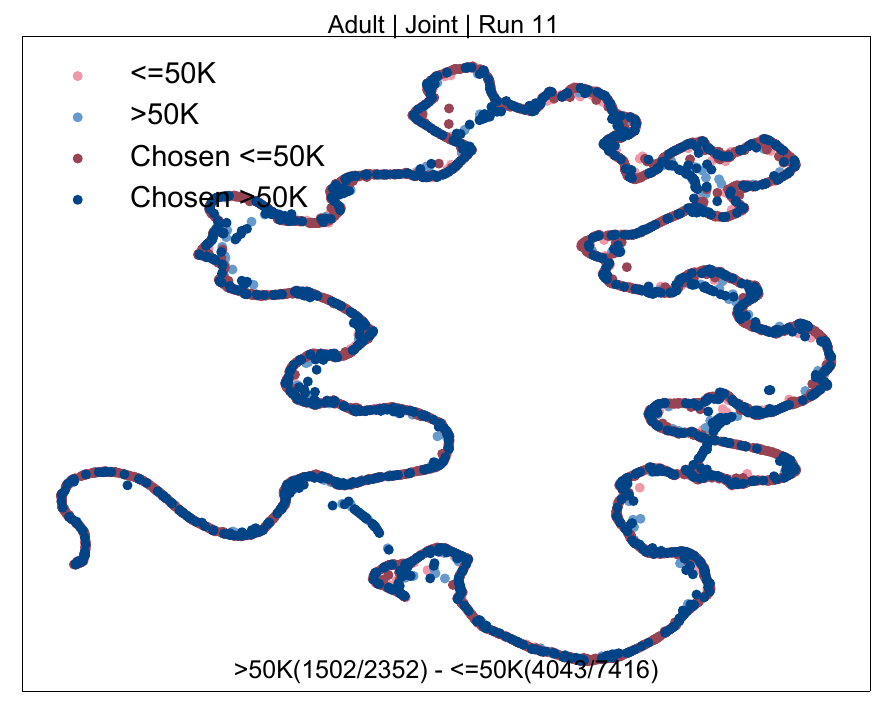}
         \caption{ }
         \label{supfig:umap_joint_adult}
     \end{subfigure}
\caption{\textbf{Impact of joint bias on the UMAP latent space.} 
Samples selected by joint bias, highlighted on the respective latent UMAP space of the labeled train set for each of the 11 datasets: (a) wine, (b) mushroom, (c) fire, (d) breast cancer, (e) pumpkin, (f) MNIST, (g) spam, (h) raisin, (i) rice, (j) pistachio, and (k) adult. Results are shown for run 11 (arbitrarily chosen).
} 
\label{supfig:umap_joint_all}
\end{figure}

\clearpage
\subsection{Semi-supervised methods on other bias induction techniques}
\begin{figure}[ht!]
\centering
\includegraphics[width=\linewidth]{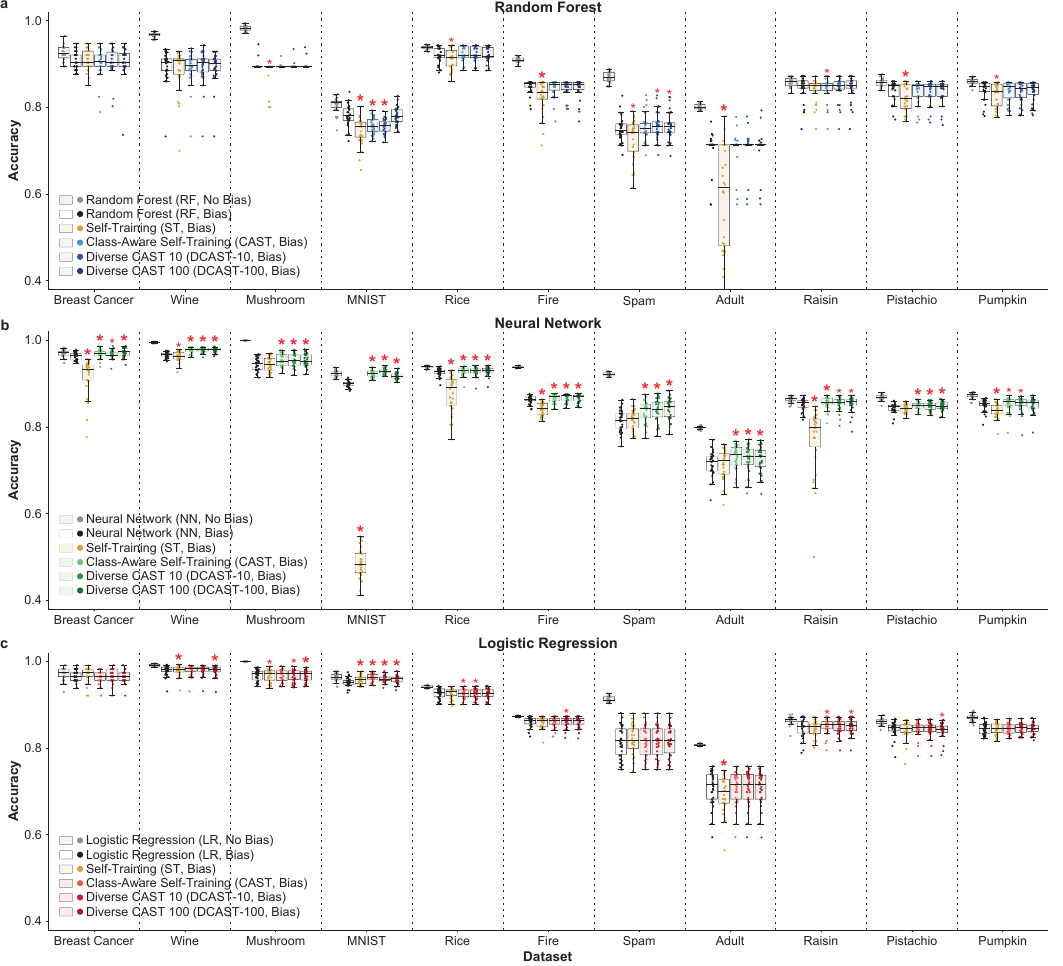}
\caption{\textbf{Performance of semi-supervised bias mitigation upon random subsampling. }
Accuracy of supervised and semi-supervised learning methods with \textbf{(a)} RF, \textbf{(b)} NN, and \textbf{(c)} LR models across 11 datasets. Results for 30 runs: each training on a different split of the train set into labeled and unlabeled sets, all evaluated on the same original test set. Models included (top to bottom): supervised RF/NN/LR models trained on the original (No Bias) or biased (Bias) labeled set; and semi-supervised RF/NN/LR models, using conventional self-training (ST) on the biased labeled train set plus the unlabeled test set, or (D)CAST on the biased labeled train set plus the unlabeled train set. Red asterisks (*) denote statistically significant changes in accuracy over 30 runs for each semi-supervised approach compared to supervised learning on the biased labeled set, using one-sided Wilcoxon signed-rank tests (larger asterisks indicate $p < 0.01$, smaller asterisks $0.01 < p < 0.05$).
}
\label{supfig:ssl_performances_random}
\end{figure}

\begin{figure}[ht!]
\centering
\includegraphics[width=\linewidth]{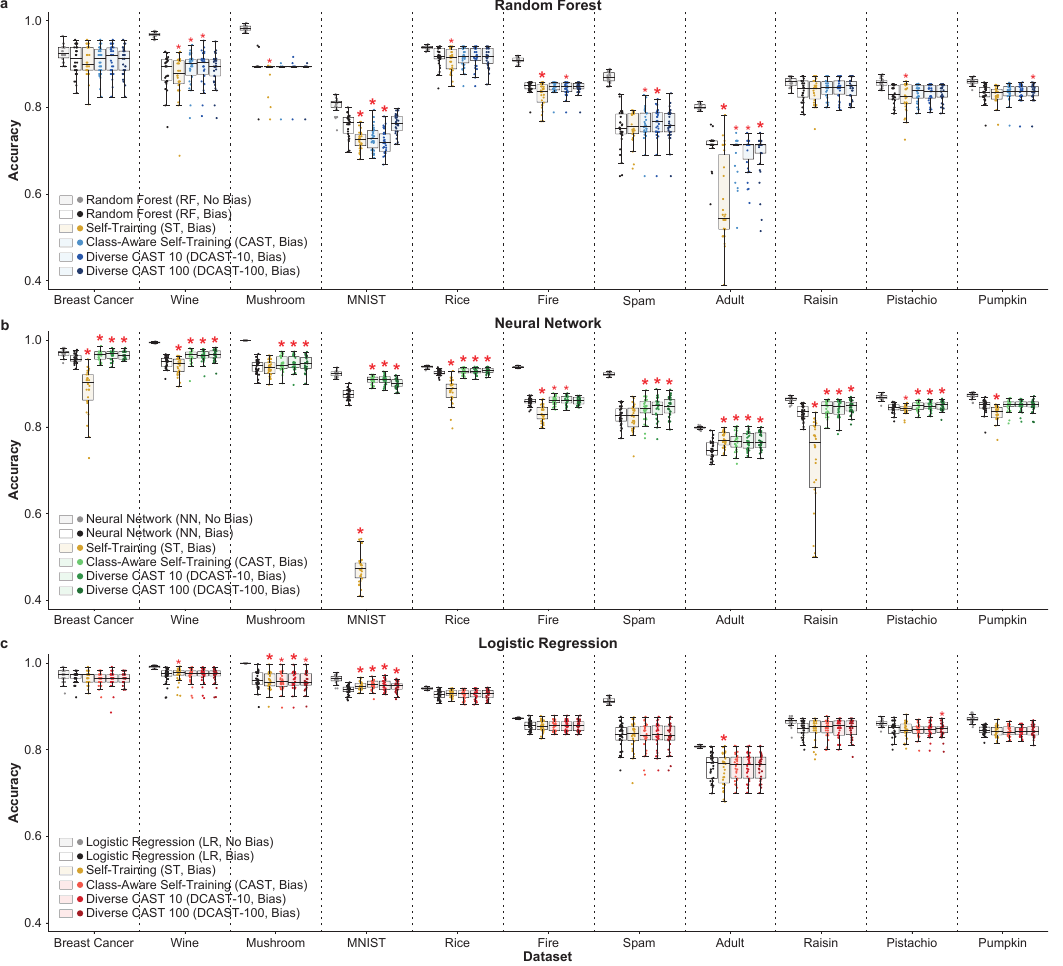}
\caption{\textbf{Performance of semi-supervised bias mitigation under Dirichlet bias. 
}
Accuracy of supervised and semi-supervised learning methods with \textbf{(a)} RF, \textbf{(b)} NN, and \textbf{(c)} LR models across 11 datasets. Results for 30 runs: each training on a different split of the train set into labeled and unlabeled sets, all evaluated on the same original test set. Models included (top to bottom): supervised RF/NN/LR models trained on the original (No Bias) or biased (Bias) labeled set; and semi-supervised RF/NN/LR models, using conventional self-training (ST) on the biased labeled train set plus the unlabeled test set, or (D)CAST on the biased labeled train set plus the unlabeled train set. Red asterisks (*) denote statistically significant changes in accuracy over 30 runs for each semi-supervised approach compared to supervised learning on the biased labeled set, using one-sided Wilcoxon signed-rank tests (larger asterisks indicate $p < 0.01$, smaller asterisks $0.01 < p < 0.05$).
}
\label{supfig:ssl_performances_dirichlet}
\end{figure}

\begin{figure}[ht!]
\centering
\includegraphics[width=\linewidth]{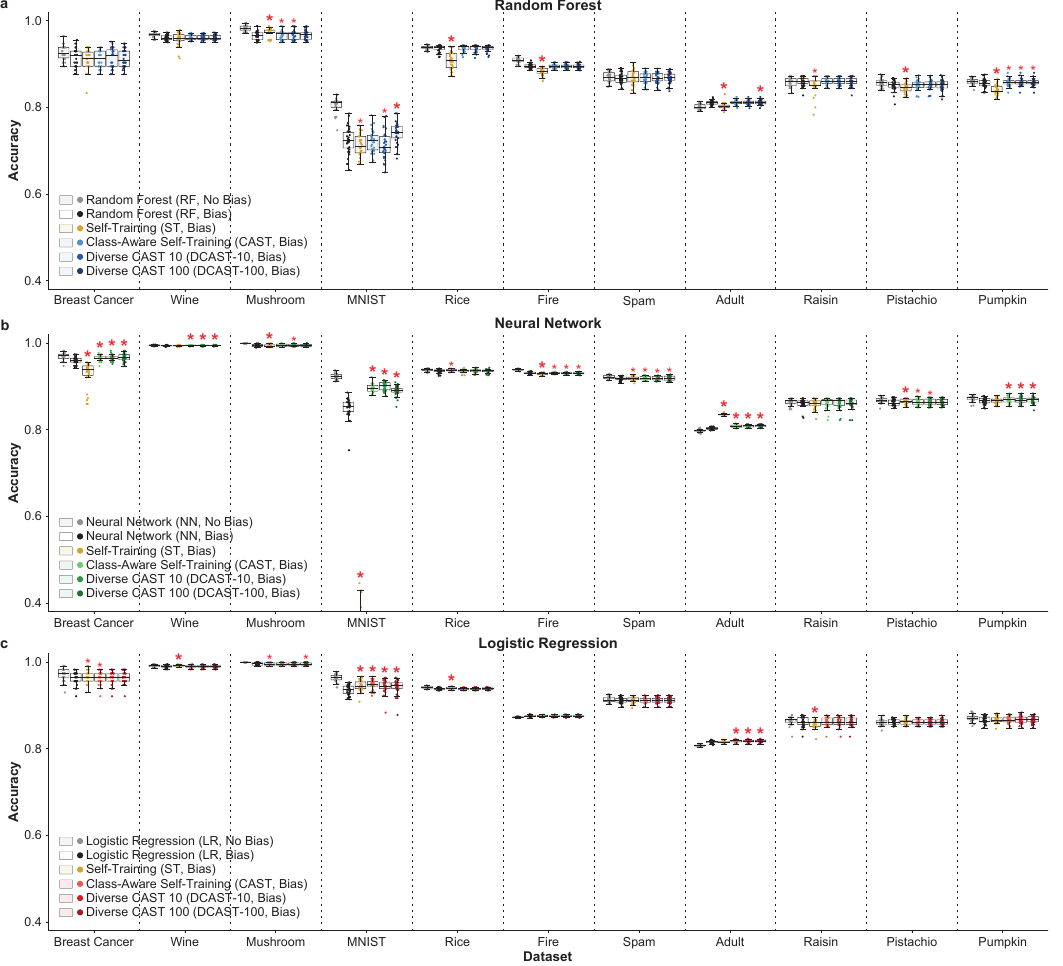}
\caption{\textbf{Performance of semi-supervised bias induction under joint bias. }
Accuracy of supervised and semi-supervised learning methods with \textbf{(a)} RF, \textbf{(b)} NN, and \textbf{(c)} LR models across 11 datasets. Results for 30 runs: each training on a different split of the train set into labeled and unlabeled sets, all evaluated on the same original test set. Models included (top to bottom): supervised RF/NN/LR models trained on the original (No Bias) or biased (Bias) labeled set; and semi-supervised RF/NN/LR models, using conventional self-training (ST) on the biased labeled train set plus the unlabeled test set, or (D)CAST on the biased labeled train set plus the unlabeled train set. 
Red asterisks (*) denote statistically significant changes in accuracy over 30 runs for each semi-supervised approach compared to supervised learning on the biased labeled set, using one-sided Wilcoxon signed-rank tests (larger asterisks indicate $p < 0.01$, smaller asterisks $0.01 < p < 0.05$).
}
\label{supfig:ssl_performances_joint}
\end{figure}

\clearpage
\section{Supplementary Tables}
\subsection{Dataset Statistics}

\clearpage
\begin{table}[!htbp]
\centering
\caption{\textbf{Class balance of original and biased labeled train sets, as well as biased selection ratio, for 11 datasets over 30 train runs.} 
The ``class balance'' columns indicate the ratio between the number of samples in the first class (Class 0) and the total number of samples in the original and the biased labeled sets. The ``Selection ratio'' columns refer to the ratio between the number of samples in the biased labeled set and the number of samples in the original labeled set per class. The columns ``Avg.'' and ``SD'' contain the average and standard deviation of the values over 30 runs, respectively. For MNIST, only the statistics for the first 2 classes are reported.
}
\begin{tabular}{ll|c|cc|cc|cc}
\textbf{Dataset} & \textbf{Bias} &
  \textbf{\begin{tabular}[c]{@{}c@{}}Original\\class\\balance\end{tabular}} &
  \multicolumn{2}{c|}{\textbf{\begin{tabular}[c]{@{}c@{}}Biased\\class\\balance\end{tabular}}}
   &
  \multicolumn{2}{c|}{\textbf{\begin{tabular}[c]{@{}c@{}}Selection\\ratio\\(class 0)\end{tabular}}}
   &
  \multicolumn{2}{c}{\textbf{\begin{tabular}[c]{@{}c@{}}Selection\\ratio\\(class 1)\end{tabular}}}
    \\ \midrule
                       &                                  &  & \textbf{Avg.} & \textbf{SD} & \textbf{Avg.} & \textbf{SD} & \textbf{Avg} & \textbf{SD} \\ \midrule
\textbf{Adult}         & \textbf{Dirichlet}       & \multirow{4}{*}{0.241} & 0.249 & 0.058 & 0.006 & 0.001 & 0.006 & 0.000 \\
\textbf{Adult}         & \textbf{Hierarchy (0.9)} &  & 0.500 & 0.000 & 0.013 & 0.000 & 0.004 & 0.000 \\
\textbf{Adult}         & \textbf{Joint}           &  & 0.271 & 0.004 & 0.653 & 0.010 & 0.557 & 0.006 \\
\textbf{Adult}         & \textbf{Random}          &  & 0.500 & 0.000 & 0.013 & 0.000 & 0.004 & 0.000 \\ \midrule
\textbf{Breast Cancer} & \textbf{Dirichlet}       & \multirow{4}{*}{0.625} & 0.478 & 0.063 & 0.337 & 0.045 & 0.614 & 0.075 \\
\textbf{Breast Cancer} & \textbf{Hierarchy (0.9)} &  & 0.505 & 0.011 & 0.353 & 0.000 & 0.576 & 0.024 \\
\textbf{Breast Cancer} & \textbf{Joint}           &  & 0.671 & 0.029 & 0.688 & 0.040 & 0.563 & 0.067 \\
\textbf{Breast Cancer} & \textbf{Random}          &  & 0.500 & 0.000 & 0.353 & 0.000 & 0.588 & 0.000 \\ \midrule
\textbf{Fire}          & \textbf{Dirichlet}       & \multirow{4}{*}{0.498} & 0.533 & 0.061 & 0.015 & 0.002 & 0.013 & 0.002 \\
\textbf{Fire}          & \textbf{Hierarchy (0.9)} & & 0.500 & 0.000 & 0.014 & 0.000 & 0.014 & 0.000 \\
\textbf{Fire}          & \textbf{Joint}           & & 0.500 & 0.008 & 0.503 & 0.015 & 0.498 & 0.014 \\
\textbf{Fire}          & \textbf{Random}          & & 0.500 & 0.000 & 0.014 & 0.000 & 0.014 & 0.000 \\ \midrule
\textbf{MNIST}         & \textbf{Dirichlet}       & \multirow{4}{*}{0.102} & 0.514 & 0.067 & 0.704 & 0.121 & 0.678 & 0.110 \\
\textbf{MNIST}         & \textbf{Hierarchy (0.9)} & & 0.496 & 0.019 & 0.641 & 0.034 & 0.665 & 0.035 \\
\textbf{MNIST}         & \textbf{Joint}           & & 0.443 & 0.054 & 0.463 & 0.087 & 0.595 & 0.096 \\
\textbf{MNIST}         & \textbf{Random}          & & 0.500 & 0.000 & 0.682 & 0.000 & 0.698 & 0.000 \\ \midrule
\textbf{Mushroom}      & \textbf{Dirichlet}       & \multirow{4}{*}{0.518} & 0.550 & 0.069 & 0.033 & 0.004 & 0.029 & 0.004 \\
\textbf{Mushroom}      & \textbf{Hierarchy (0.9)} & & 0.500 & 0.000 & 0.030 & 0.000 & 0.032 & 0.000 \\
\textbf{Mushroom}      & \textbf{Joint}           & & 0.544 & 0.011 & 0.597 & 0.015 & 0.536 & 0.014 \\
\textbf{Mushroom}      & \textbf{Random}          & & 0.500 & 0.000 & 0.030 & 0.000 & 0.032 & 0.000 \\ \midrule
\textbf{Pistachio}     & \textbf{Dirichlet}       & \multirow{4}{*}{0.427} & 0.456 & 0.065 & 0.124 & 0.018 & 0.111 & 0.013 \\
\textbf{Pistachio}     & \textbf{Hierarchy (0.9)} & & 0.500 & 0.000 & 0.136 & 0.000 & 0.102 & 0.000 \\
\textbf{Pistachio}     & \textbf{Joint}           & & 0.411 & 0.018 & 0.656 & 0.057 & 0.701 & 0.032 \\
\textbf{Pistachio}     & \textbf{Random}          & & 0.500 & 0.000 & 0.136 & 0.000 & 0.102 & 0.000 \\ \midrule
\textbf{Pumpkin}       & \textbf{Dirichlet}       & \multirow{4}{*}{0.480} & 0.504 & 0.085 & 0.105 & 0.018 & 0.095 & 0.016 \\
\textbf{Pumpkin}       & \textbf{Hierarchy (0.9)} & & 0.500 & 0.000 & 0.104 & 0.000 & 0.096 & 0.000 \\
\textbf{Pumpkin}       & \textbf{Joint}           & & 0.441 & 0.016 & 0.580 & 0.040 & 0.679 & 0.028 \\
\textbf{Pumpkin}       & \textbf{Random}          & & 0.500 & 0.000 & 0.104 & 0.000 & 0.096 & 0.000 \\ \midrule
\textbf{Raisin}        & \textbf{Dirichlet}       & \multirow{4}{*}{0.500} & 0.638 & 0.060 & 0.355 & 0.034 & 0.201 & 0.034 \\
\textbf{Raisin}        & \textbf{Hierarchy (0.9)} & & 0.500 & 0.000 & 0.278 & 0.000 & 0.278 & 0.000 \\
\textbf{Raisin}        & \textbf{Joint}           & & 0.482 & 0.024 & 0.689 & 0.042 & 0.741 & 0.045 \\
\textbf{Raisin}        & \textbf{Random}          & & 0.500 & 0.000 & 0.278 & 0.000 & 0.278 & 0.000 \\ \midrule
\textbf{Rice}          & \textbf{Dirichlet}       & \multirow{4}{*}{0.572} & 0.522 & 0.066 & 0.060 & 0.008 & 0.073 & 0.010 \\
\textbf{Rice}          & \textbf{Hierarchy (0.9)} & & 0.500 & 0.000 & 0.057 & 0.000 & 0.077 & 0.000 \\
\textbf{Rice}          & \textbf{Joint}           & & 0.604 & 0.009 & 0.630 & 0.022 & 0.553 & 0.023 \\
\textbf{Rice}          & \textbf{Random}          & & 0.500 & 0.000 & 0.057 & 0.000 & 0.077 & 0.000 \\ \midrule
\textbf{Spam}          & \textbf{Dirichlet}       & \multirow{4}{*}{0.394} & 0.524 & 0.050 & 0.072 & 0.007 & 0.043 & 0.004 \\
\textbf{Spam}          & \textbf{Hierarchy (0.9)} & & 0.500 & 0.000 & 0.069 & 0.000 & 0.045 & 0.000 \\
\textbf{Spam}          & \textbf{Joint}           & & 0.388 & 0.007 & 0.780 & 0.020 & 0.800 & 0.012 \\
\textbf{Spam}          & \textbf{Random}          & & 0.500 & 0.000 & 0.069 & 0.000 & 0.045 & 0.000 \\ \midrule
\textbf{Wine}          & \textbf{Dirichlet}       & \multirow{4}{*}{0.754} & 0.801 & 0.055 & 0.041 & 0.003 & 0.031 & 0.009 \\
\textbf{Wine}          & \textbf{Hierarchy (0.9)} & & 0.500 & 0.000 & 0.026 & 0.000 & 0.078 & 0.000 \\
\textbf{Wine}          & \textbf{Joint}           & & 0.786 & 0.007 & 0.725 & 0.043 & 0.606 & 0.048 \\
\textbf{Wine}          & \textbf{Random}          & & 0.500 & 0.000 & 0.026 & 0.000 & 0.078 & 0.000
\end{tabular}
\label{suptab:datasets_bias_stats}
\end{table}

\clearpage
\begin{table}[!htbp]
\centering
\caption{\textbf{Dataset statistics.} Statistics of the datasets used to evaluate bias induction and bias mitigation strategies.}
\begin{tabular}{lccccc}
\toprule
\textbf{Dataset} & \textbf{\begin{tabular}[c]{@{}c@{}}Number\\of\\samples\end{tabular}} & \textbf{\begin{tabular}[c]{@{}c@{}}Feature\\types\end{tabular}} & \textbf{\begin{tabular}[c]{@{}c@{}}Number\\of\\features\end{tabular}} & \textbf{Classes}                                      & \textbf{Balance} \\
\midrule
\begin{tabular}[c]{@{}c@{}}Breast\\cancer\end{tabular} & 569   & Continuous  & 30 & Malignant (0), Benign (1)          & 0: 37\%, 1: 63\% \\
Wine          & 6497  & Continuous  & 11 & Red (0), White (1)                 & 0: 25\%, 1: 75\% \\
Mushroom      & 8124  & Categorical & 22 & Poisonous (0), Edible (1)          & 0: 48\%, 1: 52\% \\
MNIST         & 1797  & Continuous  & 64 & 10 classes: 0 to 9                 & 0, ..., 9: 10\% \\
Fire          & 17442 & Mixed (1C)  & 6  & Non-extinction(0),  Extinction (1) & 0: 50\%, 1: 50\% \\ Spam          & 4601  & Continuous  & 57 & Safe (0), Spam (1)                 & 0: 61\%, 1: 39\% \\
Adult         & 48842 & Mixed (7C)  & 13 & Earns \textless{}50k (0), Earns \textgreater{}50k (1) & 0: 76\%, 1: 24\%              \\
Rice          & 3810  & Continuous  & 7  & Cammeo (0), Osmancik (1)           & 0: 43\%, 1: 57\% \\
Raisin        & 900   & Continuous  & 7  & Kecimen (0), Besni (1)             & 0: 50\%, 1: 50\% \\
Pistachio     & 2148  & Continuous  & 17 & Kirmizi (0), Siit (1)              & 0: 57\%, 1: 43\% \\
Pumpkin       & 2500  & Continuous  & 13 & Cercevelik (0), Urgup (1)          & 0: 52\%, 1: 48\%
\end{tabular}
\label{suptab:datasets}
\end{table}

\clearpage
\subsection{Hyperparameters}

\begin{table}[ht!]
    \centering
    \caption{\textbf{Parameter values of models trained using supervised learning and bias mitigation methods.} Model types: LR, logistic regression (scikit-learn implementation); RF, random forest (LightGBM implementation); and NN, neural network (Keras implementation).}
\begin{tabular}{c|l|l|c}
\textbf{Method} &
  \multicolumn{1}{l|}{\textbf{Hyperparameter}} &
  \multicolumn{1}{l|}{\textbf{Brief description}} &
  \multicolumn{1}{c}{\textbf{Value}} \\ \hline
\multirow{3}{*}{\textbf{LR}} & penalty & Regularization type & L2 \\ \cline{2-4}
 & C & Inverse of reg. strength & 5.0 \\ \cline{2-4}
 & max\_iter & Maximum iteration & 100 \\ \hline
\multirow{8}{*}{\textbf{RF}} & subsample & \begin{tabular}[c]{@{}l@{}}Subsample ratio of \\ training samples\end{tabular} &
  0.9 \\ \cline{2-4}
  & subsample\_freq & Frequency of subsample & 1 \\ \cline{2-4}
 & min\_child\_weight & \begin{tabular}[c]{@{}l@{}}Minimum sum of instance weight\\ (Hessian) needed in a child\end{tabular} & 0.01 \\ \cline{2-4}
 & reg\_lambda & L2 regularization term & 5 \\ \cline{2-4}
 & num\_leaves & Maximum tree leaves & 31 \\ \cline{2-4}
 & max\_depth & Maximum tree depth  & -1 \\ \cline{2-4}
 & n\_estimators & Number of decision trees & 100 \\ \hline
\multirow{3}{*}{\textbf{NN}} & activation & Activation function of layers & RELU \\ \cline{2-4}
 & optimize & Optimization strategy & Adam \\ \cline{2-4}
 & loss & Loss function to optimize & Cross entropy \\
\end{tabular}
\label{suptab:hyperparams}
\end{table}

\newpage
\clearpage